\documentclass{article}

\usepackage{microtype}
\usepackage{graphicx}
\usepackage{subfigure}
\usepackage{units}
\usepackage{multirow}
\usepackage{amsfonts}
\usepackage{booktabs} 
\usepackage[figuresright]{rotating}
\usepackage{amsmath,amsfonts,amssymb}
\usepackage{bbm}
\usepackage{bm}
\usepackage[dvipsnames]{xcolor}
\usepackage{color, colortbl}
\definecolor{Gray}{gray}{0.9}
\definecolor{babypink}{rgb}{0.96, 0.76, 0.76}
\definecolor{LemonChiffon}{RGB}{255, 250, 205}
\definecolor{Thistle}{RGB}{255, 225, 255}
\definecolor{LightBlue1}{RGB}{178, 223, 238}
\definecolor{LightBlue}{RGB}{191, 239, 255}
\definecolor{RoyalBlue3}{RGB}{0, 0, 205}
\definecolor{codegreen}{rgb}{0,0.6,0}
\definecolor{codegray}{RGB}{71, 69, 84} 
\definecolor{codepurple}{rgb}{0.58,0,0.82}
\definecolor{backcolour}{rgb}{0.95,0.95,0.92}

\usepackage{caption}
\usepackage{listings}
\lstdefinestyle{mystyle}{
  backgroundcolor=\color{backcolour}, commentstyle=\color{codegreen},
  keywordstyle=\color{magenta},
  numberstyle=\tiny\color{codegray},
  stringstyle=\color{codepurple},
  basicstyle=\ttfamily\footnotesize,
  breakatwhitespace=false,         
  breaklines=true,                 
  captionpos=b,                    
  keepspaces=true,                 
  numbers=left,                    
  numbersep=8pt,                  
  showspaces=false,                
  showstringspaces=false,
  showtabs=false,                  
  tabsize=2
}
\usepackage{hyperref}

\usepackage[accepted]{icml2021}

\usepackage{amsmath}
\usepackage{amssymb}
\usepackage{mathtools}
\usepackage{amsthm}
\usepackage{enumitem}

\usepackage[capitalize,noabbrev]{cleveref}

\newtheorem{theorem}{Theorem}[section]
\newtheorem{proposition}{Proposition}

\theoremstyle{definition}
\newtheorem{definition}{Definition}
\newtheorem{assumption}{Assumption}

\icmltitlerunning{Group \& Reweight: A Novel Cost-Sensitive Approach to Mitigating Class Imbalance in Network Traffic Classification}

\begin{document}

\twocolumn[
\icmltitle{
Group \& Reweight: A Novel Cost-Sensitive Approach to Mitigating Class Imbalance in Network Traffic Classification}



\icmlsetsymbol{equal}{*}
\icmlsetsymbol{correspondence}{*}

\begin{icmlauthorlist}
\icmlauthor{Wumei Du}{nudt}
\icmlauthor{Dong Liang}{correspondence,nudt}
\icmlauthor{Yiqin Lv}{nudt}
\icmlauthor{Xingxing Liang}{nudt2}
\icmlauthor{Guanlin Wu}{cod}
\icmlauthor{Qi Wang}{nudt}
\icmlauthor{Zheng Xie}{correspondence,nudt}
\end{icmlauthorlist}

\icmlaffiliation{nudt}{College of Science, National University of Defense Technology, Changsha, China}
\icmlaffiliation{nudt2}{College of Systems Engineering, National University of Defense Technology, Changsha, China}
\icmlaffiliation{cod}{Academy of Military Science, Beijing, China}

\icmlcorrespondingauthor{Dong Liang and Zheng Xie}{dongliangnudt@nudt.edu.cn;xiezheng81@nudt.edu.cn}

\icmlkeywords{Machine Learning, ICML}

\vskip 0.3in
]



\printAffiliationsAndNotice{}  
\begin{abstract}
Internet services have led to the eruption of network traffic, and machine learning on these Internet data has become an indispensable tool, especially when the application is risk-sensitive.
This paper focuses on network traffic classification in the presence of severe class imbalance.
Such a distributional trait mostly drifts the optimal decision boundary and results in an unsatisfactory solution.
This raises safety concerns in the network traffic field when previous class imbalance methods hardly deal with numerous minority malicious classes.
To alleviate these effects, we design a \textit{group \& reweight} strategy for alleviating class imbalance.
Inspired by the group distributionally optimization framework, our approach heuristically clusters classes into groups, iteratively updates the non-parametric weights for separate classes, and optimizes the learning model by minimizing reweighted losses.
We theoretically interpret the optimization process from a Stackelberg game and perform extensive experiments on typical benchmarks.
Results show that our approach can not only suppress the negative effect of class imbalance but also improve the comprehensive performance in prediction.
\end{abstract}

\begin{figure}[ht]
\begin{center}
\centerline{\includegraphics[width=0.45\textwidth]{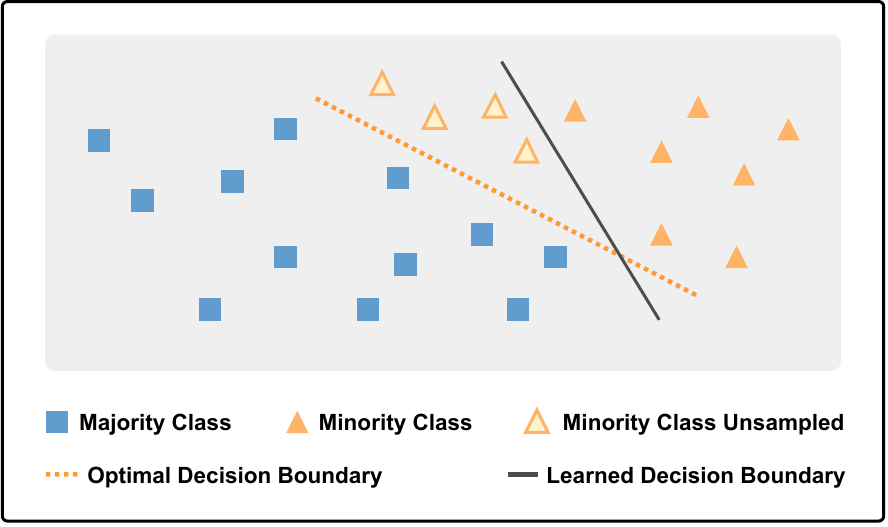}}
\caption{\textbf{Drift of the decision boundary in binary classification.}
Due to the nature of the class imbalance, the machine learning obtained decision boundary tends to deviate from the optimal one.
}
\vspace{-30pt}
\label{class_imbalance_effect}
\end{center}
\end{figure} 

\section{Introduction}
The past few decades have witnessed a surge in Internet traffic due to the quick development of the Internet of Things (IoTs) \citep{rose2015internet} and the great demand for data resources for training machine learning models.
The potential value of increasing Internet traffic also brings many concerns, and security is the most crucial factor in Internet services.
To prevent hacker attacks, malware, phishing, etc., and protect users' privacy, the technical staff encapsulates most network traffic according to encryption protocols before it is transmitted.

Considering the inherent risk sensitivity to applications, there has been growing increasing interest in analyzing these encrypted traffic data with the help of machine learning methods \citep{zhao2021network}.
Particularly, network traffic classification has played an essential role in network security, privacy protection, network management, and intrusion detection \citep{nguyen2008survey}.
Through in-depth research and development of relevant technologies and methods, we can effectively deal with the challenges brought by encrypted traffic and enhance network security.

\textbf{Existing Challenges:}
When it comes to network traffic data classification, there emerge several challenges.
Unlike general class imbalance learning problems \citep{LANGO2022116962,TSAI2024111267}, the classes of traffic data are mostly numerous, and there exhibits a high imbalance among classes.
Particularly, as illustrated in Fig. \ref{Data_quantity}, the minority but sensitive classes, such as malicious traffic, tend to occupy a small proportion of the whole dataset.
These traits in statistics cause the prevalent machine learning models to suffer from the \textit{decision boundary drift} in the literature \citep{wang2017novel,douzas2018effective,huang2023neural}.
As illustrated in Fig. \ref{class_imbalance_effect}, examples of the minority class near the oracle boundary can be easily identified as the majority class by the classifier due to the boundary drift.
To alleviate such an effect, researchers have developed a collection of strategies, such as data augmentation, under/over-sampling, cost-sensitive learning, and so forth.
Even so, obtaining a robust classification model in a simple yet effective way is still challenging.
Among existing techniques, data augmentation methods and cost-sensitive learning methods enjoy special popularity.
The family of data augmentation relies on sampling strategies, generative models, or other mix-up tricks \citep{elreedy2019comprehensive,chou2020remix}.
Such a family either requires a particular configuration in sampling ratios \citep{shelke2017review} or can easily produce unnatural instances \citep{sampath2021survey}.
Benefiting from diverse available mechanisms, the cost-sensitive family is more flexible in configurations \citep{madabushi2019cost}, nevertheless, these mechanisms demand several manual or complicated adjustments in practice.
Meanwhile, \textit{the existence of too many classes in the network traffic makes it challenging for previous class imbalance methods to handle}.

\textbf{Research Motivations \& Methods:}
In the presence of numerous network traffic classes exhibiting significant imbalances, traditional imbalance learning techniques often struggle to identify the minority classes that have very few instances effectively.
Thus, new algorithms need to be developed to mitigate the negative effects of degraded classification performance caused by extreme class imbalance.
This work stresses the importance of the class grouping mechanism in optimization, which is also inspired by group distributionally robust optimization \citep{sagawa2019distributionally,rahimian2022frameworks}.
Technically, we establish connections between distributional robustness and cost-sensitive strategies and present a novel approach to resolving the network traffic class imbalance issue.
In circumventing the complicated mechanism design, we propose \textit{a group \& reweight strategy for processing the imbalance problem with numerous classes}, \textit{interpret the operations from the Stackelberg game}, and \textit{conduct theoretical analysis on optimization}.
In detail, we at first cluster classes into distinct groups based on the validated results of a proxy neural network that was initially trained in a standard supervised manner, as shown in Fig. \ref{grouping}.
Then we dynamically assign weights to their loss functions at a group level.
The key insight is that classification results over the minority classes can be viewed as the worst cases in optimization, and we can take the intermediate results into optimization.
Such a design can \textit{deal with the network traffic dataset of extremely many classes} and \textit{stably schedule the cost-sensitive class weights} for performance balance.

\textbf{Primary Contributions \& Outlines:}
The remainder is organized as follows.
Section \ref{sec:related_work} summarizes the literature work on traffic data classification.
In Section \ref{sec:preliminary}, we introduce mathematical notations and necessary preliminaries.
Then, we present the developed approach in Section \ref{sec:dr_loss}.
The experimental results and analysis are reported in Section \ref{sec:exp_analysis}, followed by the conclusion and future extension.
To sum up, our contribution is three-fold:
\begin{enumerate}
    \item We propose a novel approach for network traffic classification, which offers a heuristic and effective rule to group numerous traffic classes for distributionally robust optimization to suppress the negative effect of class imbalance.
    \item The optimization process in our approach can be interpreted as approximately resolving a Stackelberg game, and we constitute the equilibrium concept as the robust solution in network traffic classification and conduct theoretical analysis.
    \item The empirical results show that, in network traffic classification, our approach is superior to existing state-of-the-art (SOTA) baselines for minority classes, while achieving comparable performance for majority classes.
\end{enumerate}

\section{Literature Review}\label{sec:related_work}
\textbf{Background Knowledge of Encrypted Traffic.}
Specifically, the sender converts the plaintext data packet into ciphertext through an encryption algorithm and then packages the ciphertext into an encrypted data packet for transmission according to the encryption protocol. 
Only when the receiver has the corresponding decryption algorithm can the encrypted data packet be decrypted and processed, thereby effectively protecting the data packet's confidentiality, integrity, and security and preventing it from being detected, attacked, and tampered with during network transmission. 
As the computing capabilities of computers, mobile terminals, and other devices continue to improve, various encryption technologies and encryption protocols are constantly updated, resulting in the rapid growth of encrypted traffic data. 
At the same time, encryption technology provides opportunities for many unscrupulous cyber criminals. 
Various types of malware and cyber attacks also use encryption to avoid detection by various network defense security systems. 
Network traffic classification is generally the prerequisite for network intrusion detection and an essential task in network security. 

\textbf{Categories on Traffic Class Imbalance Learning Methods.}
To alleviate the effect of class imbalance, researchers have developed a collection of strategies, which can be mainly divided into three categories:

(i) Under/Over-sampling aims to reduce the size of the majority classes and increase that of the minority classes respectively in the learning process, forcing almost equal quantities relative to each other \citep{park2021combined,zuech2021detecting,chen2024adaptive,li2023oaldpc}.
Random Under Sampling (RUS) and Random over sampling (ROS) are two classical approaches.
RUS randomly removes some majority classes' samples which might lead to information loss \citep{lotfollahi2020deep}, while ROS produces copies for minority classes to achieve balance \citep{NK2024112186}.
Synthetic Minority Over-sampling Technique (SMOTE) \citep{chawla2002smote, yan2018new,zhang2020effective} creates synthetic samples rather than copies, which is proven to be effective in imbalanced network traffic classification, but it tends to introduce noise and causes instability in learning \citep{guo2021combating}.

(ii) Data augmentation focuses on density estimation and sample generative mechanism.
By leveraging existing dataset, the generative artificial intelligence models \citep{wang2023large} can sample more synthetic samples for the minority class \citep{zavrak2020anomaly,guo2021combating,lin2022machine,monshizadeh2021improving,khanam2022towards,d2021network,baarli2021and,xu2020toward}.
Several studies report impressive performance based on deep generative models for data augmentation, especially the Generative Adversarial Network (GAN) \citep{goodfellow2014generative, thi2025comparative}.
Despite the potential of GAN-based data augmentation methods \citep{guo2021combating,wang2020packetcgan,ding2022imbalanced,KHAN2024122778}, the generative process sometimes produces samples from the low-density region and results in unsatisfactory minority ones.
Meanwhile, generative models are challenging in stable training and suffer from mode collapse, which sometimes leads to the lack of diversity of generated samples.

(iii) The mechanism of cost-sensitive learning is to reweight the costs of various classes in the training process \citep{rezvani2023broad}.
The role of class-specific weights is to penalize the case when the classification of the minority is severely under performance in the optimization process.
Some researchers integrated the class-specific costs with the loss function for deep learning classification on imbalanced datasets \citep{wang2016training,gupta2022cse}.
The earlier work that introduced cost-sensitive learning to handle the class imbalance problem in network security is in intrusion detection for IoTs \citep{telikani2021cost}.
Subsequently, some deep learning frameworks that adopt this strategy have been developed for network traffic classification on unbalanced traffic data, such as support vector machine\citep{dong2021multi}, and recurrent neural networks \citep{zhu2023cmtsnn}.
Most of these cases are heuristic in class weight assignment.
In view of the flexibility in configurations of the cost-sensitive learning, we try to trade off the overall performance and
computational efficiency by dynamically assigning cost weights at a group level.
This study can also be viewed as cost-sensitive learning.
Additionally, it is the first study to assign cost weights through clustering, which distinguishes it from previous works.

\section{Preliminaries}\label{sec:preliminary}
\textbf{Notations.}
Let $\mathbf{x}\in\mathcal{X}\subseteq\mathbb{R}^{d}$ be the vector of each traffic data as the input for classification.
We denote the prediction label for each input feature by $y\in\mathcal{Y}=\{1,2,\dots,\mathcal{C}\}$.
The parameter of the classification model is expressed as $\bm\theta\in\bm\Theta$.
We write the distribution of data points as $p(\mathbf{x},y)$.
The dataset of interest is $D=\{(\mathbf{x}_i,y_i)\}_{i=1}^{N}$, where $N$ is the number of training samples.
We partition the dataset into $D=\cup_{c=1}^{\mathcal{C}}D_{c}$, where $D_{c}$ denotes the data points that belong to the class $c$.

The goal in network traffic classification is to learn a map $f_{\bm\theta}:\mathcal{X}\mapsto\mathcal{Y}$ that tries to minimize the classification errors, e.g., $\mathbb{E}_{p(\mathbf{x},y)}\left[\mathbbm{1}{[f_{\bm\theta}(\mathbf{x})\neq y]}\right]$, as much as possible.
The following will detail the classification model and fundamental knowledge about the Stackelberg game, which are prerequisites for our developed method.

\subsection{Multiclass Imbalance Learning}
We employ deep learning models for multiclass imbalanced network traffic classification.
The fully connected deep neural networks (DNNs) are constructed as the backbone for generality to enable multiclass classification.
In mathematics, the DNN consists of the feature extraction module $\mathbf{\Phi}=\{\mathbf{W}^{i},\mathbf{b}^{i}\}_{i=1}^{H}$ and the output module $\mathbf{\Phi}_{o}=\{\mathbf{W}_{c}^{o},b_{c}^{o}\}_{c=1}^{\mathcal{C}}$.
Hence, the estimated probability of a data point $(\mathbf{x},y)$ for all classes can be computed as:
\begin{equation}\label{output_classification_prob}
    p(\hat{y}=c\vert\mathbf{x};\bm\theta)=\frac{\exp\Big({\mathbf{\Phi}(\mathbf{x})}^{\intercal}\mathbf{W}_{c}^{o}+b_{c}^{o}\Big)}{\sum_{i=1}^{\mathcal{C}}\exp\Big({\mathbf{\Phi}(\mathbf{x})}^{\intercal}\mathbf{W}_{i}^{o}+b_{i}^{o}\Big)},
\end{equation}
where the model parameter is $\bm\theta:=\{\mathbf{\Phi},\mathbf{\Phi}_{o}\}$ and $\hat{y}$ is the predicted result.

Furthermore, we can introduce $\mathcal{L}(D;\bm\theta)$ the negative log-likelihood (NLL) as the loss function in multiclass classification with the help of \textbf{Eq.} (\ref{output_classification_prob}):
\begin{equation}
    \begin{split}\label{output_classification_nll}
        \min_{\bm\theta\in\bm\Theta}\mathcal{L}(D;\bm\theta):=\sum_{i=1}^{N}\sum_{c=1}^{\mathcal{C}}
        -\mathbbm{1}{[y_{i}=c]}\ln p(\hat{y}_{i}=c\vert\mathbf{x}_{i};\bm\theta).
    \end{split}
\end{equation}

\subsection{Stackelberg Game}

The two-player Stackelberg game is a strategic interaction model frequently utilized in game theory. 
In such a game, two competitive players engage in a sequential decision-making process to maximize their utility functions.
The leader player $\mathcal{P}_L$ makes decisions first, followed by the follower player, referred to as $\mathcal{P}_F$, who makes decisions subsequently after accessing the leader's decisions.
This work focuses on a two-player zero-sum Stackelberg game, which can be mathematically depicted as $\mathcal{SG}:=\langle\mathcal{P}_L,\mathcal{P}_F; \{{\bm \eta}\in \Delta_{\mathcal{C}}\}, \{{\bm \theta} \in \bm\Theta \}; \mathcal{F}({\bm \eta}, {\bm \theta})\rangle$.
$\Delta_{\mathcal{C}}$ (resp. $\bm\Theta$) represents the action space for player $\mathcal{P}_L$ (resp. player $\mathcal{P}_F$), and $\mathcal{F}({\bm \eta}, {\bm \theta})$ (resp., $-\mathcal{F}({\bm \eta}, {\bm \theta}) $) represents player $ \mathcal{P}_L $'s (resp., player $\mathcal{P}_F$'s) utility function.

Central to the Stackelberg game is the concept of strategic leadership, whereby the leader's actions influence the follower's subsequent decisions. 
Being aware of this influence, the leader strategically chooses its actions to maximize its utility while considering the follower's response. 
Conversely, recognizing the leader's choices, the follower optimizes its decisions accordingly.

Thus, the equilibrium of $\mathcal{SG}$ can be solved by {\emph{backward induction}}, that is,
\begin{subequations}\label{main_sgame_eq}
    \begin{align}
        &\max_{{\bm \eta}\in \Delta_{\mathcal{C}}}
        \mathcal{F}({\bm \eta},{\bm \theta}_{*}({\bm \eta}))
        \quad\text{s.t.}
        \
        {\bm \theta}_{*}({\bm \eta})=\arg\min_{{\bm \theta}\in \bm\Theta}\mathcal{F}({\bm \eta},{\bm \theta})\tag{\text{\textbf{3.a:} Leader's Decision-Making}}
        \\
        &\min_{{{\bm \theta}\in\bm\Theta}}\mathcal{F}({\bm \eta}, {\bm \theta})\tag{\text{\textbf{3.b:} Follower's Decision-Making}}.
    \end{align}
\end{subequations}

Note that the constraint in \textbf{Eq.} (\ref{main_sgame_eq}.a) is called the leader player $\mathcal{P}_L$'s {\emph{best response function}} in game theory.
Such a sequential structure inherently offers the leader a strategic advantage, as it can anticipate and preempt the follower's reactions. 
Consequently, the Stackelberg game is characterized by an asymmetric power dynamic between the leader and follower, where the leader's strategic precommitment shapes the subsequent interaction between the players.

\section{Group Distributionally Robust Class Imbalance Learner}\label{sec:dr_loss}

As previously mentioned, the existence of class imbalance tends to result in decision boundary drift, which can be risky or even lead to catastrophic results in network traffic classification scenarios.
One plausible way to alleviate this effect is to penalize the classification errors via adjusting weights to individual classes \citep{chou2020remix}.
Nevertheless, effectively enabling weight assignment in training processes remains a tricky problem, and there is a limited mechanism for explaining the process.
To this end, we introduce \underline{\textbf{G}}roup \underline{\textbf{D}}istributionally \underline{\textbf{R}}obust \underline{\textbf{C}}lass \underline{\textbf{I}}mbalance \underline{\textbf{L}}earner (\texttt{\textbf{GDR-CIL}}) as a competitive candidate method to \textit{group} and \textit{reweight} network traffic class examples in optimization.
Please refer to Algorithm \ref{alg: go4balance} for details.

\subsection{Distributional Robustness}

This work reduces the decision boundary drift in class imbalance learning to the distributional robustness issue \citep{sagawa2019distributionally,rahimian2022frameworks}.
Robustness is a crucial consideration in tail risk cases \citep{shen-etal-2024-assessing}.
Here, we consider the robustness regarding the classification performance of the minority classes.

To this end, we can translate the optimization problem in the presence of class imbalance as follows:
\begin{equation}
\begin{split}
\label{optimization:minmax}\max_{{\bm \eta}\in \Delta_{\mathcal{C}}}\min_{\bm\theta\in\bm\Theta}\mathcal{F}({\bm \eta},\bm\theta) := \mathbb{E}_{p(D)}\Big[{\bm \eta}^{\intercal}\bm{\mathcal{L}}_{\mathcal{C}}(D;\bm\theta)\Big]+\lambda\mathcal{R}(\bm \eta),
\end{split}
\end{equation}
where the loss term $\bm{\mathcal{L}}_{\mathcal{C}}(D;\bm\theta)=\left[\ell(D_1;\bm\theta),\dots,\ell(D_{\mathcal{C}};\bm\theta)\right]^{\intercal}$ denotes the vector of losses for each class.
Moreover, ${\bm \eta}$ corresponds to an element in $\Delta_{\mathcal{C}}$ which is the $(\mathcal{C}-1)$-dimensional simplex and the term $\mathcal{R}(\bm \eta)$ works as a regularization to avoid the collapse of $\bm \eta$ into vertices.

The max-min optimization problem in \textbf{Eq.} \eqref{optimization:minmax} can be understood as a two-player zero-sum Stackelberg game \citep{breton1988sequential,li2017review}.
The leader $ \mathcal{P}_L $ adjusts the weights for different classes over the probability simplex aiming to maximize $\mathcal{F}({\bm \eta}, \bm\theta
)$.
The minimization operator executes in the parameter
space, corresponding to the follower $ \mathcal{P}_F $ in $\mathcal{SG}$ with the utility function $-\mathcal{F}({\bm \eta}, \bm\theta)$.

\subsection{Group \& Reweight Surrogate Loss Functions}\label{sec:construction of loss}

Unlike previous methods, this work reduces class imbalance learning to a distributionally robust optimization at a class group level \citep{sagawa2019distributionally}.
We make a hypothesis that multi-classes retain unobservable group structures in losses, and placing various weights over a collection of groups in optimization can balance the performance of all classes.
To this end, our approach executes the following training pipeline to obtain robust classifiers:
\begin{enumerate}
    \item Cluster classes into groups with the help of validation performance based on the trained proxy classifier;
    \item Evaluate the classifier's performance on these groups based on updated model parameters;
    \item Respectively update the groups' weights and model parameters from learning losses;
    \item The steps (2)/(3) loop until reaching the convergence.  
\end{enumerate}
Next, we report details on how to construct the distributionally robust surrogate loss function and schedule the group weights.

\textbf{Grouping classes in the imbalanced dataset.}
First, we initially train the previously mentioned backbone model with the collected dataset in a standard supervised way, namely without any class imbalance suppression mechanism.
Such a proxy training phase derives a proxy classifier for empirically assessing the class imbalance circumstance.
Then, based on the proxy training results on the validation dataset, we adopt the clustering strategy as follows: (i) treat classes with nearly zero \textit{F1-score}s as separate groups and (ii) cluster the remainder classes into other limited groups with the instance numbers for each class as input.
Such a design pays more attention to hardly discriminating minority classes.
Also, note that the results for the majority classes with similar training samples are close to each other in terms of initial training performance.
In light of this, we leverage empirical assessment and observations as prior knowledge to define groups based on the training data heuristically.

\begin{figure*}[h]
\centerline{\includegraphics[width=1.0\textwidth]{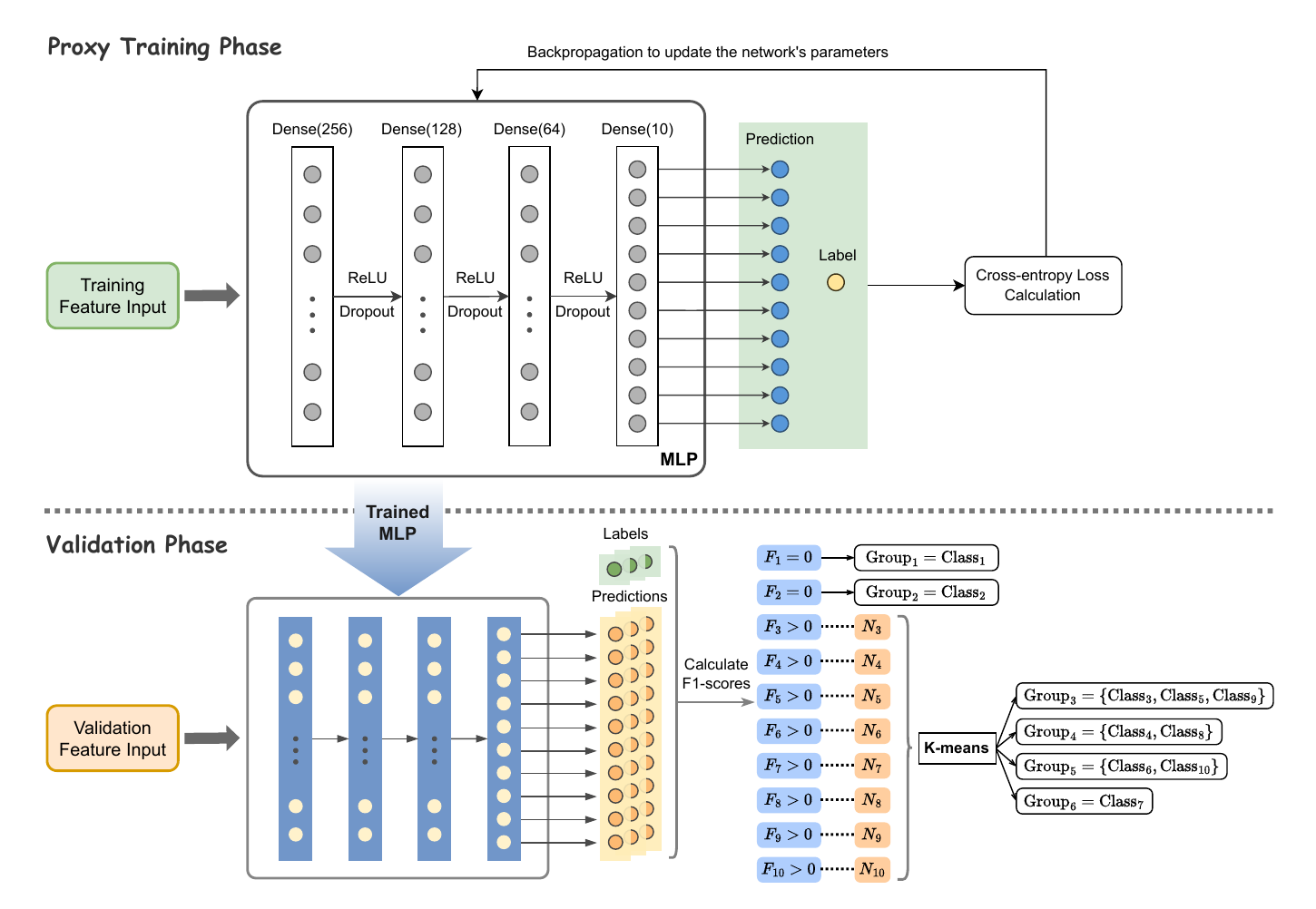}}
\caption{\textbf{The flow chart of grouping mechanism.}
We provide a toy example of grouping in the imbalance traffic dataset with ten classes.
The grouping mechanism is implemented before the formal classification training.
It involves the proxy training phase and the validation phase.
We first train an MLP in a standard supervised manner as the proxy training phase. 
Then, we calculate the F1-scores for the ten classes in the validation phase, denoted as $F_1, \dots, F_{10}$ respectively.
Given $F_1$ and $F_2$ 0 and others non-zero, classes 1 and 2 are assigned to groups 1 and 2.
For classes $\{3,4,.\dots,10\}$, we employ the K-means algorithm to cluster them into distinct groups based on their instance counts in the training dataset, represented by $N_3, N_4, \ldots, N_{10}$ in this example.
}
\label{grouping}
\end{figure*} 

Here, we provide a flow chart of the grouping mechanism as illustrated in Fig. \ref{grouping}.
We first train a multi-layer perceptron (MLP) network on the training dataset, assuming that it contains ten classes, and then validate the trained MLP on the validation dataset.
Classes 1 and 2 are set as separate groups because their \textit{F1-score}s are both equal to $0$. 
For classes 3 through 10, we adopt a data clustering algorithm, such as the K-means \citep{lloyd1982least}, to group them based on similar training sample numbers.
The role of initial training is to serve the formal training as getting groups fixed.
With the grouping mechanism applied, the training data comprises $(\bm{\mathbf{x}},y,g)$ triplets, where $g\in\mathcal{G}=\{1,2,\dots, K\}$ represents the group label for each input feature $\mathbf{x}$ and $K$ is denoted as the number of groups which varies depending on the dataset.

\textbf{Group distributionally robust loss function.}
The distribution of data points can be written as $p(\mathbf{x},y,g)$.
Given an integer $K$ as the number of groups, we denote the assignment matrix by $\mathbf{\Gamma}\in\mathbb{R}^{K\times\mathcal{C}}$, in which $\gamma_{k,c}$ equals $1$ if the $k$-th group contains the $c$-th class and $0$ otherwise.
The group number generally is smaller than the class number, \textit{i.e.}, ${1}<{K}\leq{\mathcal{C}}$.
Thus, the loss for each group can be formulated by:
\begin{equation}
\label{loss_for_groups}
\bm{\mathcal{L}}_{\mathcal{G}}(D;\bm\theta)=\mathbf{\Gamma}\bm{\mathcal{L}}_{\mathcal{C}}(D;\bm\theta),
\end{equation}
where $\bm{\mathcal{L}}_{\mathcal{G}}(D;\bm\theta) = [\ell_g(D_1;{\bm \theta}), \dots, \ell_g(D_K;{\bm \theta})]^{\intercal} \in\mathbb{R}^{K}$ is the loss vector at the group level, and its $k$-th element corresponds to the summation of intra-group losses $\sum_{c=1}^{\mathcal{C}}\gamma_{k,c}\ell(D_{c};\bm\theta)$.
Naturally, we expect the classification model to generalize well on average and the worst-case group.
However, there are variations in the generalization gaps across groups, which result in the gap between average and worst-group test performances.
To bridge the gap, we attempt to reduce training loss in the groups with larger generalization gaps.
Concretely, we adopt the operation in \citep{cao2019learning,sagawa2019distributionally} to obtain $\mathbf{v}_{\mathcal{G}}=\left[\nicefrac{B}{\sqrt{n_1}},\dots,\nicefrac{B}{\sqrt{n_K}}\right]^{\intercal}=\left[v_1,\dots,v_K\right]^{\intercal}$, where $n_k$ is the group size for the $k$-th group and $B$ the model calibration constant treated as a hyper-parameter.
The scaling with $\nicefrac{1}{\sqrt{n_k}}$ reflects that small groups are more likely to overfit than large groups.
By applying the group calibrations, the model is inclined to pay more attention to fitting the smaller groups.

\textbf{Iteratively estimating the groups' weights.}
We produce groups' weights with the grouping mechanism described above.
To balance different groups, we place the weights over groups ${\bm{\omega}}=\left[\omega_{1},\dots,\omega_{K}\right]^{\intercal}\in\Delta_{K}$, where $\omega_{k}\geq{0}$ is specific to the $k$-th group.
In the proposed approach, intra-group classes share the same weight, while different groups possess various weights.
That is, $ \eta_i = \eta_j = \omega_k $ when the $i$-th and $j$-th classes belong to the $k$-th group.
Thus, the minimax optimization problem in \textbf{Eq.} \eqref{optimization:minmax} is further regularized as follow:
\begin{equation}
\begin{split}
\label{optimization:minmax_adjust}\max_{\bm\omega\in \Delta_K}\min_{\bm\theta\in\bm\Theta}\mathcal{F}({\bm \omega}, \bm\theta) := \mathbb{E}_{p(D)}\Big\{{\bm \omega}^{\intercal}\Big[\mathbf{\Gamma}\bm{\mathcal{L}}_{\mathcal{C}}(D;\bm\theta)+\mathbf{v}_{\mathcal{G}}\Big]\Big\}.
\end{split}
\end{equation}

The optimization problem in \textbf{Eq.} \eqref{optimization:minmax_adjust} can be interpreted as approximately solving a stochastic two-player zero-sum Stackelberg game $\mathcal{SG} = \langle\mathcal{P}_L,\mathcal{P}_F; \{{\bm \omega} \in \Delta_{K}\}, \{{\bm \theta}\in \bm\Theta\};\mathcal{F}({\bm \omega},{\bm \theta})\rangle $.
The two players sequentially compete to maximize separate utility functions in $\mathcal{SG}$, which can be described as
\begin{subequations}
\begin{align}
\mathcal{SG}:
        {\bm \omega}^{(t)}=
        \underbrace{
        \arg\max_{{\bm \omega}\in \Delta_K}~ {\mathbb{E}_{p(D)}\Big \{{\bm \omega}^{\intercal}\left[\mathbf{\Gamma}\bm{\mathcal{L}}_{\mathcal{C}}(D;\bm\theta^{(t)})+\mathbf{v}_{\mathcal{G}}\right]\Big \}}}_{\text{Leader Player}},
        \label{Leader_Player}\\
        \bm\theta^{(t+1)}=
    \underbrace{
        \arg\min_{\bm\theta\in\bm\Theta}~
        {\mathbb{E}_{p(D)}\Big \{({\bm \omega}^{(t)})^{\intercal}\left[\mathbf{\Gamma}\bm{\mathcal{L}}_{\mathcal{C}}(D;\bm\theta)+\mathbf{v}_{\mathcal{G}}\right]\Big \}}}_{\text{Follower Player}}.
        \label{Follower_Player}    
\end{align}
\end{subequations}

As for approximately obtaining the above best responses, 
the update of $\bm\omega$ takes the following rule the same as that in \citep{sagawa2019distributionally}.
Let ${\bm \omega}^{(t)}$ be the vector of groups' weights under the model's parameters $\bm\theta^{(t)}$, then the $i$-th element in $\bm\omega$ is updated in the $t$-th iteration by:
\begin{align}
    \tag*{}\omega_{i}&^{(t)}=\frac{\omega_{i}^{(t-1)}\exp\Big(\beta (\ell_g(D_{i};\bm\theta^{(t)})+v_{i})\Big)}{\sum_{k=1}^{K}\omega_{k}^{(t-1)} \exp\Big(\beta(\ell_g(D_{k};\bm\theta^{(t)})+v_{k})\Big)}, \\
    \label{Calibrate_group_weights}&\forall i\in\{1,\dots,K\}
    \
    \text{and}
    \
    {\bm \omega}^{(t-1)}=\Big[\omega_{1}^{(t-1)},\dots,\omega_{K}^{(t-1)}\Big]^{\intercal},
\end{align}
where $\ell_g(D_{k};\bm\theta^{(t)})$ is the last time evaluated performance for the class $k$, and $\sum_{k=1}^{K}\omega_{k}^{(t-1)}=1$.
$\beta$ denotes the temperature hyper-parameter to control the scale of the loss value.
For the minority classes, their loss are greater and therefore the weights assigned to them should be higher.
For expositional clarity, define $\omega_{i}^{(t)} = \mathcal{H}_i({\bm\omega}^{(t-1)}, {\bm \theta}^{(t)}), \forall i\in\{1,\dots,K\},$ and $ \mathcal{H}({\bm\omega}^{(t-1)}, {\bm \theta}^{(t)}) = [\mathcal{H}_1({\bm\omega}^{(t-1)}, {\bm \theta}^{(t)}), \dots, \mathcal{H}_K({\bm\omega}^{(t-1)}, {\bm \theta}^{(t)})]^{\intercal} $.

For the model parameters' update, we apply the stochastic gradient descent to $\bm\theta$ with the learning rate $\epsilon$ and the sampled batch sample $D\sim p(D)$:
\begin{equation}\label{SGD}
    \begin{split}
        \bm\theta^{(t+1)}=\bm\theta^{(t)}-\epsilon\cdot\nabla_{\bm\theta}{\bm \omega}^{(t)\intercal}\mathbf{\Gamma}\bm{\mathcal{L}}_{\mathcal{C}}(D;\bm\theta^{(t)}).
    \end{split}
\end{equation}

\subsection{Solution Concepts in \texttt{\textbf{GDR-CIL}}}

Next, we focus on another crucial question in this part:
{\bf \textit{What is the notion of the convergence point in network traffic classification}?}
In response to this question, we need to formulate the corresponding solution concept in $\mathcal{SG}$. Here, we induce the global Stackelberg equilibrium as the solution concept.
\begin{definition}[Global Stackelberg Equilibrium]\label{def:equilibrium}
    Let $({\bm{\omega}}_{*},{\bm \theta}_{*})\in\Delta_K\times\bm\Theta$ be the solution. 
    With the leader ${\bm{\omega}}_{*}\in \Delta_K $ and the follower ${\bm \theta}_{*}\in\bm\Theta$, $({\bm{\omega}}_{*},{\bm \theta}_{*})$ is called a \textit{global Stackelberg equilibrium} if the following inequalities are satisfied, $\forall {\bm \omega}\in\Delta_K$ and $\forall {\bm \theta} \in\bm\Theta$,
    \begin{equation*}
        \inf_{{\bm \theta}' \in \bm\Theta} \mathcal{F}({\bm \omega}, {\bm \theta}') \le \mathcal{F}({\bm \omega}_{*},{\bm{\theta}}_{*}) \le \mathcal{F}({\bm \omega}_{*},{\bm{\theta}}).
    \end{equation*}
\end{definition}

\begin{proposition}[Existence of Equilibrium]\label{prop:equilibrium}
   Suppose that $\bm\Theta$ is compact, the global Stackelberg equilibrium of $\mathcal{F}({\bm \omega}, {\bm \theta})$ always exists.
\end{proposition}

\textit{\textbf{Proof.}} 
Since $\mathcal{F}({\bm \omega}, {\bm \theta})$ is continuous with respect to ${\bm \omega}$ and $ {\bm \theta} $, along with that $ \Delta_K \in \mathbb{R}^{K} $ and $ \bm\Theta $ are compact, Proposition \ref{prop:equilibrium} holds due to the extreme-value theorem.
$\hfill\blacksquare$

We can further explain the obtained equilibrium $({\bm{\omega}}_{*},{\bm \theta}_{*})$ in a more intuitive way.
That is, the leader has already specified the optimal strategy ${\bm{\omega}}_{*}$ for reweighting different groups,
while the follower cannot find another model parameter except ${\bm \theta}_{*}$ to suppress the negative effect of class imbalance in the worst groups.
As a result, we can obtain a robust class imbalance learner from the devised optimization strategies.

\textbf{Convergence Analysis.} While the existence of a global Stackelberg equilibrium is guaranteed, finding it using existing optimization techniques is NP-hard. Following the approach in \citep{fiez2020implicit}, we focus instead on the local Stackelberg equilibrium, as defined in Definition \ref{local_SE} below.
\begin{definition}[Local Stackelberg Equilibrium]\label{local_SE}
    Let $({\bm\omega}^{*},{\bm \theta}^{*})\in \Delta_K \times{\bm \Theta}$ be the solution. 
    With the leader ${\bm\omega}^{*}\in\Delta_K$ and the follower ${\bm \theta}^{*}\in{\bm \Theta}$,
    $({\bm\omega}^{*},{\bm \theta}^{*})$ is called a \textit{local Stackelberg equilibrium} for the leader if the following inequalities hold, $\forall {\bm\omega}\in {\bm \Omega}^\prime \subset \Delta_K $,
    $$\inf_{{\bm\theta}\in\mathcal{S}_{{\bm\Theta}^{\prime}}({\bm\omega}^{*})}\mathcal{F}({\bm\omega}^{*},{\bm \theta})\geq\inf_{{\bm \theta}\in\mathcal{S}_{{\bm \Theta}^{\prime}}({\bm \omega})}\mathcal{F}({\bm \omega},{\bm \theta}), $$ 
    $\text{where}\ \mathcal{S}_{{\bm\Theta}^{\prime}}({\bm\omega}):=\{\bar{\bm \theta}\in{\bm\Theta}^{\prime}\vert\mathcal{F}({\bm \omega},\bar{\bm \theta})\leq\mathcal{F}({\bm \omega},{\bm \theta}),\forall{\bm \theta}\in{\bm \Theta}^{\prime}\}.$
\end{definition}

To analyze the convergence of local Stackelberg equilibrium, we first present the following assumptions regarding the loss function $\bm{\mathcal{L}}_{\mathcal{G}}(D;\bm\theta)$ and the weight update function $ {\mathcal{H}({\bm \omega}, \bm\theta)} $.
\begin{assumption}\label{asum: loss}
    $\bm{\mathcal{L}}_{\mathcal{G}}(D;\bm\theta) $ is $\eta$-Lipschitz continuous \textit{w.r.t.} ${\bm \theta}$ for any dataset $D$.
\end{assumption}
\begin{assumption}\label{asum: weight}
    There exist two constants $\xi > 0$ and $ 0<\gamma<1 $ such that $\Vert \nabla_{\bm \theta}{\mathcal{H}({\bm \omega}, \bm\theta)}\Vert_2 \le \xi $ and $\Vert \nabla_{\bm \omega}{\mathcal{H}({\bm \omega}, \bm\theta)}\Vert_2 \le \gamma $.
\end{assumption}

We can now derive the following convergence rate theorem for the second player.
\begin{theorem}[Convergence Rate for the Second Player]\label{them:cr}
\textit{Let the iteration sequence in optimization be: $\cdots\mapsto\{{\bm \omega}^{(t-1)},{\bm \theta}^{(t)}\}\mapsto\{{\bm \omega}^{(t)},{\bm \theta}^{(t+1)}\}\mapsto\cdots\mapsto\{{\bm \omega}^*,{\bm \theta}^*\}$, with the converged equilibrium $({\bm \omega}^*,{\bm \theta}^*)$.
Under the Assumptions \ref{asum: loss}, \ref{asum: weight}, and supposing that $||{\bf I} -\epsilon\nabla^{2}_{\bm \theta \bm \theta}\mathcal{F}({\bm \omega}^*;{\bm \theta}^*)||_2<1-\epsilon\xi\eta$, we can have $\lim_{t\to\infty}\frac{||{\bm \theta}^{(t+1)}-{\bm \theta}^*||_2}{||{\bm \theta}^{(t)}-{\bm \theta}^*||_2}\leq 1$, and the iteration converges with the rate $\left(||{\bf I} -\epsilon\nabla^{2}_{\bm \theta \bm \theta}\mathcal{F}({\bm \omega}^*;{\bm \theta}^*)||_2+\epsilon\xi\eta\right)$.}
\end{theorem}

For the sake of brevity, the proof of Theorem \ref{them:cr} is deferred to the Appendix \ref{proof:cr}.

\begin{algorithm}[t!]
    \caption{\textbf{Training Processes in \texttt{\textbf{GDR-CIL}} Methods.}}
    \begin{algorithmic}[1]
        \STATE \textbf{Input}:  
        Maximum iteration number $T$;
        Batch size ${N_{\rm b}}$;
        Learning rate $\epsilon$; 
        Temperature hyper-parameter $\beta$;
        Group number $K$;
        Group size for the $K$ groups $n_1, \cdots, n_K$;
        Assignment matrix $\mathbf{\Gamma}$;
        Calibration hyper-parameter $B$;
        Class number $\mathcal{C}$;
        Weight vector $\bm{\omega}$.
        
        \STATE Initialize model parameters $\bm\theta^{(1)}$;
        \STATE Initialize each group weight ${\omega_k}^{(0)}\leftarrow \frac{1}{K}$;
        \STATE Compute the calibration vector
        $\mathbf{v}_{\mathcal{G}}=\left[\nicefrac{B}{\sqrt{n_1}},\dots,\nicefrac{B}{\sqrt{n_K}}\right]^{\intercal}$;

        \FOR{$t = 1 : T$}
            \STATE Randomly sample a batch of training samples: 
            $D^{(t)}:=\{(\mathbf{x}_n,y_n,g_n)\}_{n=1}^{N_{\rm{b}}}, \text{where}~ \mathbf{x}_{n}\in\mathcal{X}, y_n\in\mathcal{Y}, g_n\in\mathcal{G};$
            \STATE Compute the empirical risk for each class at $t$-th iteration $\bm{\mathcal{L}}_{\mathcal{C}}(D^{(t)};\bm\theta^{(t)})$;
            \STATE Compute the group risk $\bm{\mathcal{L}}_{\mathcal{G}}(D^{(t)};\bm\theta^{(t)})$ in \textbf{Eq.}  (\ref{loss_for_groups}) with the help of $\mathbf{\Gamma}$;

            \STATE{\textcolor{blue}{\text{\# \textbf{\texttt{The Leader's Decision-Making}}}}}

            \STATE Estimate group weights ${\bm \omega}^{(t)}(\bm\theta^{(t)})$ according to \textbf{Eq.} (\ref{Calibrate_group_weights}), ${\bm \omega}^{(t-1)}$ and $\bm{\mathcal{L}}_{\mathcal{G}}(D^{(t)};\bm\theta^{(t)})$;

            \STATE{\textcolor{blue}{\text{\# \textbf{\texttt{The Follower's Decision-Making}}}}}
            
            \STATE Update the model parameters to obtain $\bm\theta^{(t+1)}$ by applying the stochastic gradient descent with the learning rate $\epsilon$ and $\bm\omega^{(t)}$ according to \textbf{Eq.} (\ref{SGD}).
        \ENDFOR
    \end{algorithmic}
    \label{alg: go4balance}
\end{algorithm}

{\bf Computational Complexity.} 
The computational complexity of \texttt{\textbf{GDR-CIL}} consists of two main components: (1) the clustering method used during the grouping stage, and (2) the calculation of empirical risk for each class in every iteration, where the latter has a complexity of $\mathcal{O}(N_b \mathcal{C}^2)$.
Therefore, the overall complexity of \texttt{\textbf{GDR-CIL}} depends on the specific clustering method adopted.
In the experiments presented in this paper, the K-means algorithm is employed for grouping, with a computational complexity of $\mathcal{O}(n K T')$, where $n = \sum_{i=1}^K n_i$ denotes the total number of data points, and $T'$ represents the number of iterations performed by the K-means algorithm.
In summary, the overall computational complexity of  \texttt{\textbf{GDR-CIL}} is $\mathcal{O}(n K T' + N_b \mathcal{C}^2)$.

\begin{figure*}[h]
\centerline{\includegraphics[width=0.99\textwidth]{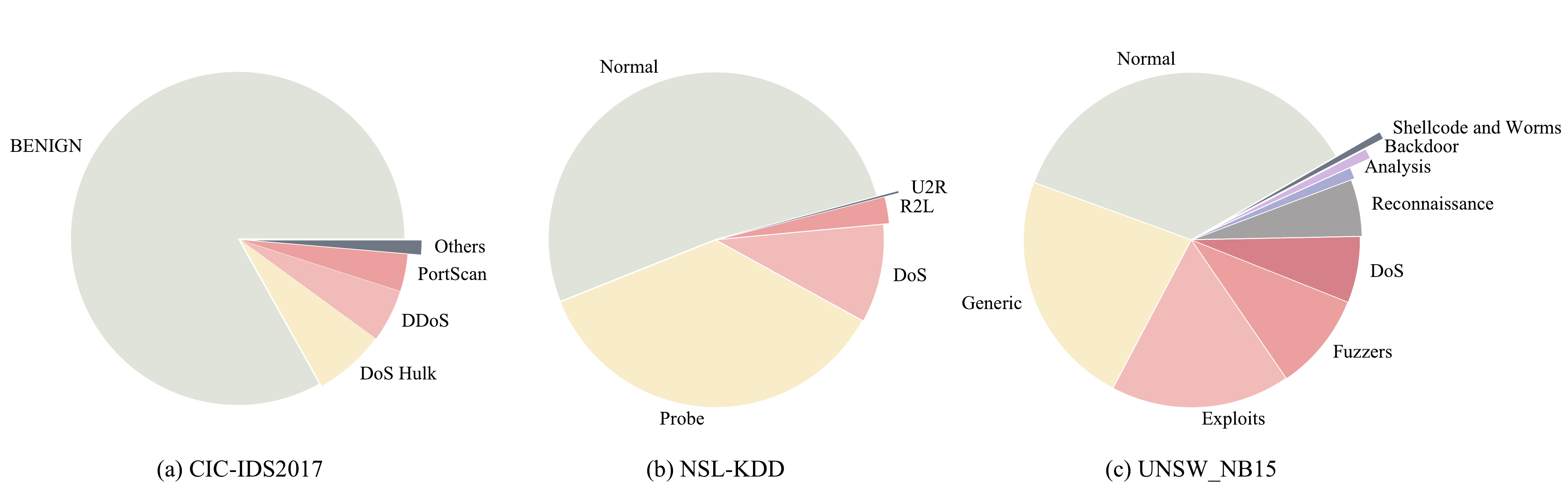}}
\caption{\textbf{Data volumes for each class of the three datasets.}
In panel (a), ``Others'' contains 11 minority classes, which are DoS GoldenEye, FTP-Patator, DoS slowloris, DoS slowhttptest, SSH-Patator, Bot, Web Attack Brute Force, Web Attack XSS, Infiltration, Web Attack Sql Injection and Heartbleed, each of them accounting for less than 0.5\% of CIC-IDS2017 dataset.
}
\label{Data_quantity}
\end{figure*}

\section{Experimental Results and Analysis}\label{sec:exp_analysis}
This section starts with the commonly-seen benchmarks and baselines in class imbalance network traffic classification, conducts extensive experiments, and performs quantitative analysis.

\subsection{Experimental Settings}\label{Experimental settings}
Here, we present the benchmarks, baselines, and evaluation metrics used in this work.

\textbf{Benchmarks:}
Three typical datasets are used to examine the performance of the proposed method, including CIC-IDS2017 \citep{sharafaldin2018toward}, NSL-KDD \citep{tavallaee2009detailed} and UNSW-NB15 datasets \citep{moustafa2015unsw}.
Please refer to Appendix \ref{Benchmarks} for details.

Fig. \ref{Data_quantity} displays the data volumes for each class of the three datasets.
It can be seen that the traffic data categories are extremely imbalanced in these datasets.



	

\textbf{Baselines:}
Deep imbalance learning is considered for network traffic classification.
Here, we compare \texttt{\textbf{GDR-CIL}} with typical and SOTA baselines achieved by various strategies.
They include expected risk minimization (ERM) \citep{vapnik1991principles}, Focal loss \citep{lin2017focal}, CB \citep{cui2019class}, Mixup-DRW \citep{zhang2021bag}, Remix-DRW \citep{chou2020remix}, LDAM-DRW \citep{cao2019learning}, and LDR-KL \citep{zhu2023label}.
ERM is the foundation of cost-sensitive learning, which is a general framework used in machine learning to find a model that minimizes the average loss.
The other six baselines, as well as our method, belong to the cost-sensitive method.
In contrast to these cost-sensitive baselines, our method is the first to implement the acquisition of prior knowledge through clustering to compute the cost weights.
We refer the reader to Appendix \ref{Baseline} for more details on these baselines.

\textbf{Evaluation Metrics:}
Four classical evaluation indicators are used to evaluate the performance of all methods in the experiment:
(1) \textit{Specificity} is the true negative rate, measuring the proportion of actual negative samples correctly predicted as negative.
(2) \textit{F1-score} takes the harmonic mean of another two performance metrics, namely \textit{Precision} and \textit{Recall}, to balance performance.
(3) \textit{G-mean} is the geometric mean of \textit{Specificity} and \textit{Recall}.
(4) \textit{AUC} refers to the area under the Receiver Operating Characteristic (ROC) curve which shows the trade-off between the True Positive Rate (TPR), also known as \textit{Recall}, and the False Positive Rate (FPR) at various decision thresholds, measuring the model's ability to distinguish between positive and negative classes.

The above comprehensive indicators are derived from four basic metrics obtained from the classification model, which are True Positive (TP), False Positive (FP), True Negative (TN), and False Negative (FN) in the confusion matrix.
This work considers multiclass network traffic classification; hence, samples from one specific traffic class are treated as positive, with others as negative in evaluation.
Here, the TP and TN represent the number of positive and negative samples that are correctly classified, respectively.
The FP indicates the number of negative samples misclassified as positive, while the FN counts the number of positive samples misclassified as negative.

We calculate \textit{Recall}, \textit{Precision} and \textit{FPR} from the basic metrics, $\textit{Recall}={\rm TP}/({\rm TP+\rm FN})$, $\textit{Precision}={\rm TP}/({\rm TP+\rm FP})$ and $\textit{FPR}={\rm FP}/({\rm FP+\rm TN})$.
The resulting comprehensive indicators are further obtained from the operations:
\begin{equation}
\begin{split}
\textit{Specificity}=\frac{\rm TN}{\rm TN+FP},\\
\end{split}
\end{equation}
\begin{equation}
\begin{split}
\textit{F1-score}=\frac{2 \times \textit{Recall} \times \textit{Precision}}{\textit{Recall}+\textit{Precision}}, \\
\end{split}
\end{equation}
\begin{equation}
\begin{split}
\textit{G-mean}=\sqrt{\textit{Recall}\times\textit{Specificity}}.\\
\end{split}
\end{equation}

\textit{AUC} can be calculated directly with the scikit-learn package in Python, which typically ranges between $0.5$ and $1.0$.
Note that these indicators \textit{F1-score}, \textit{Specificity}, \textit{G-mean}, and \textit{AUC} are always positively related to the class imbalance performance.

\begin{figure*}[h]
\centerline{\includegraphics[width=0.9\textwidth]{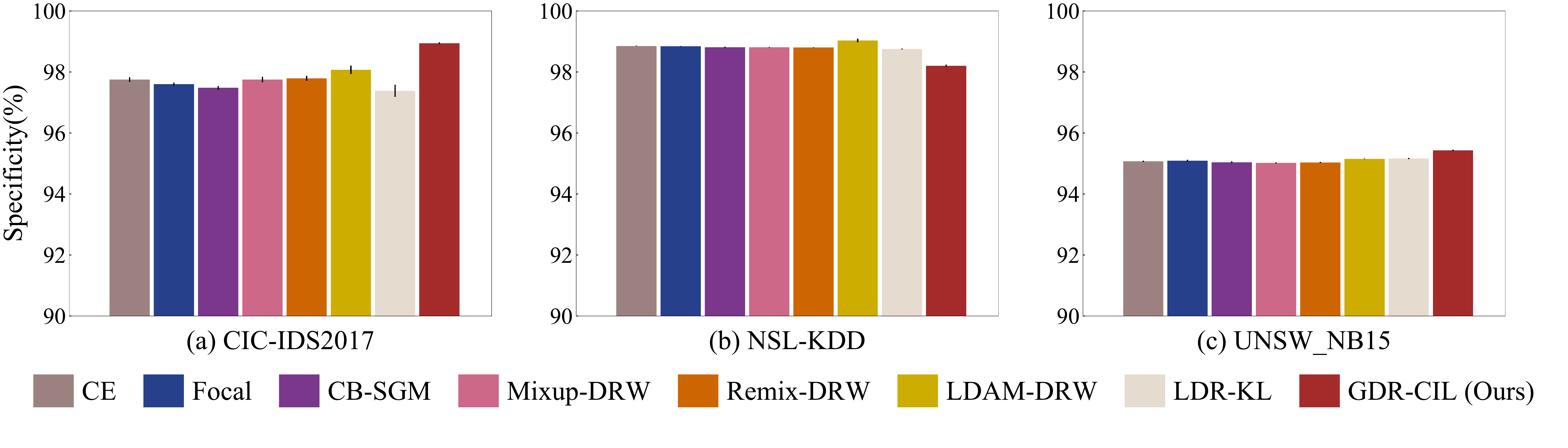}}
\caption{\textbf{The average \textit{Specificitie}s of different methods on three datasets (5 runs with error bars reported).}}
\label{Specificity-bar}
\end{figure*}

\textbf{Implementation Details:}
We randomly split the dataset into standard training-validation-testing datasets.
The validation dataset works for hyper-parameter configurations in the training process, and the testing dataset is used to evaluate experimental results.
For the sake of generality, we employ an MLP as the backbone of deep neural networks, as shown in Fig. \ref{grouping}.
It is a fully connected neural network with three hidden layers, and we use ReLU as the activation function.
We optimized the neural network using stochastic gradient descent with a learning rate of $1\mathrm{e}-2$ and a weight decay of $1\mathrm{e}-4$.
Please refer to Appendix \ref{Implementation} for details.

\subsection{Experimental Results}\label{Experimental Results}

\setlength{\tabcolsep}{3.0pt}
\begin{table*}[htbp]
	\centering
	\caption{
		\textbf{\textit{F1-score}s obtained by different methods on CIC-IDS2017 dataset (5 runs).}
        Underlined traffic categories belong to the minority class.
        The results are expressed in \% with standard deviations.}
    \vspace{0.1cm}
    \scalebox{0.84}{
	\begin{tabular}{c|ccccccc|c}
		\toprule
		Traffic Category & ERM  & {Focal} & {CB-SGM} & {Mixup-DRW} & {Remix-DRW} & {LDAM-DRW} & {LDR-KL} & {\texttt{\textbf{GDR-CIL}} (Ours)}\\
            \toprule
		BENIGN & \cellcolor{LightBlue}\textbf{97.51}\scriptsize{$\pm$\textbf{0.05} }& 97.33\scriptsize{$\pm$0.04 }& 97.32\scriptsize{$\pm$0.03 }& 97.48\scriptsize{$\pm$0.05 }& 97.47\scriptsize{$\pm$0.03 }& 97.07\scriptsize{$\pm$0.35 }& 97.39\scriptsize{$\pm$0.10 }& 94.94\scriptsize $\pm$0.06\\
            DoS Hulk & 95.62\scriptsize{$\pm$0.46 }& 94.18\scriptsize{$\pm$0.39 }& 95.03\scriptsize{$\pm$0.34 }& 95.61\scriptsize{$\pm$0.45 }& 95.59\scriptsize{$\pm$0.39 }& 94.74\scriptsize{$\pm$0.37 }& \cellcolor{LightBlue}\textbf{96.05}\scriptsize{$\pm$\textbf{0.29} }& 91.68\scriptsize$\pm$0.08\\
            DDoS & 98.50\scriptsize{$\pm$0.04 }& 98.09\scriptsize{$\pm$0.07 }& 97.55\scriptsize{$\pm$0.06 }& 98.5\scriptsize{$\pm$0.11 }& 98.46\scriptsize{$\pm$0.02 }& 98.43\scriptsize{$\pm$0.10 }& \cellcolor{LightBlue}\textbf{98.91}\scriptsize{$\pm$\textbf{0.03} }& 95.02\scriptsize$\pm$0.10\\
            PortScan & 91.42\scriptsize{$\pm$0.02 }& 91.15\scriptsize{$\pm$0.10 }& 91.31\scriptsize{$\pm$0.05 }& 91.42\scriptsize{$\pm$0.02 }& 91.44\scriptsize{$\pm$0.01 }& \cellcolor{LightBlue}\textbf{91.50}\scriptsize{$\pm$\textbf{0.04} }& 89.81\scriptsize{$\pm$0.90 }& 90.22\scriptsize$\pm$0.11\\
            \underline{{DoS GoldenEye}} & 98.61\scriptsize{$\pm$0.04 }& 98.30\scriptsize{$\pm$0.01 }& 98.40\scriptsize{$\pm$0.03 }& 98.56\scriptsize{$\pm$0.03 }& 98.48\scriptsize{$\pm$0.04 }& 96.93\scriptsize{$\pm$0.09 }& \cellcolor{LightBlue}\textbf{98.78}\scriptsize{$\pm$\textbf{0.02} }& 97.06\scriptsize$\pm$0.07\\
            \underline{{FTP-Patator}} & 97.89\scriptsize{$\pm$0.05 }& 98.11\scriptsize{$\pm$0.10 }& 97.7\scriptsize{$\pm$0.03 }& 97.85\scriptsize{$\pm$0.03 }& 97.76\scriptsize{$\pm$0.05 }& 98.11\scriptsize{$\pm$0.17 }& \cellcolor{LightBlue}\textbf{98.49}\scriptsize{$\pm$\textbf{0.10} }& 97.44\scriptsize$\pm$0.03\\
		\underline{{DoS slowloris}}& 97.03\scriptsize{$\pm$0.06 }& \cellcolor{LightBlue}\textbf{97.53}\scriptsize{$\pm$\textbf{0.12} }& 96.06\scriptsize{$\pm$0.16 }& 96.85\scriptsize{$\pm$0.08 }& 96.77\scriptsize{$\pm$0.06 }& 96.43\scriptsize{$\pm$0.29 }& 96.89\scriptsize{$\pm$0.18 }& 95.30\scriptsize$\pm$0.14\\
            \underline{{DoS Slowhttptest}} & 98.23\scriptsize{$\pm$0.02 }& \cellcolor{LightBlue}\textbf{98.34}\scriptsize{$\pm$\textbf{0.03} }& 97.70\scriptsize{$\pm$0.02 }& 98.26\scriptsize{$\pm$0.04 }& 98.26\scriptsize{$\pm$0.02 }& 98.19\scriptsize{$\pm$0.03 }& 98.24\scriptsize{$\pm$0.03 }& 97.74\scriptsize$\pm$0.05\\
            \underline{{SSH-Patator}} & 94.72\scriptsize{$\pm$0.03 }& 94.88\scriptsize{$\pm$0.24 }& 94.14\scriptsize{$\pm$0.02 }& 94.58\scriptsize{$\pm$0.09 }& 94.42\scriptsize{$\pm$0.08 }& 95.11\scriptsize{$\pm$0.22 }& 95.24\scriptsize{$\pm$0.17 }& \cellcolor{LightBlue}\textbf{95.50}\scriptsize$\pm$\textbf{0.00}\\
            \underline{{Bot}} & \cellcolor{LightBlue}\textbf{77.48}\scriptsize{$\pm$\textbf{0.06} }& 76.08\scriptsize{$\pm$0.70 }& 74.47\scriptsize{$\pm$0.05 }& 77.29\scriptsize{$\pm$0.07 }& 77.17\scriptsize{$\pm$0.20 }& 73.20\scriptsize{$\pm$3.53 }& 76.77\scriptsize{$\pm$2.30 }& 55.72\scriptsize$\pm$0.60\\
            \underline{{Web Attack Brute Force}} & 69.03\scriptsize{$\pm$0.03 }& 69.88\scriptsize{$\pm$0.46 }& 69.00\scriptsize{$\pm$0.13 }& 69.05\scriptsize{$\pm$0.03 }& 69.05\scriptsize{$\pm$0.05 }& 59.61\scriptsize{$\pm$9.50 }& \cellcolor{LightBlue}\textbf{70.44}\scriptsize{$\pm$\textbf{0.37} }& 39.08\scriptsize$\pm$2.02\\
            \underline{{Web Attack XSS}} &  0.00\scriptsize{$\pm$0.00}& 0.00\scriptsize{$\pm$0.00}& 0.00\scriptsize{$\pm$0.00}& 0.00\scriptsize{$\pm$0.00}& 0.00\scriptsize{$\pm$0.00}& 9.26\scriptsize{$\pm$7.29 }& 0.00\scriptsize{$\pm$0.00}& \cellcolor{LightBlue}\textbf{41.13}\scriptsize$\pm$\textbf{3.73}\\
		\underline{{Infiltration}} & 0.00\scriptsize{$\pm$0.00}& 0.00\scriptsize{$\pm$0.00}& 16.91\scriptsize{$\pm$4.28 }& 0.00\scriptsize{$\pm$0.00}& 0.00\scriptsize{$\pm$0.00}& 1.77\scriptsize{$\pm$1.58 }& 0.00\scriptsize{$\pm$0.00}& \cellcolor{LightBlue}\textbf{36.31}\scriptsize$\pm$\textbf{0.99}\\
            \underline{{Web Attack Sql Injection}} & 0.00\scriptsize{$\pm$0.00}& 0.00\scriptsize{$\pm$0.00}& 0.00\scriptsize{$\pm$0.00}& 0.00\scriptsize{$\pm$0.00}& 0.00\scriptsize{$\pm$0.00}& 0.19\scriptsize{$\pm$0.17 }& 0.00\scriptsize{$\pm$0.00}& \cellcolor{LightBlue}\textbf{6.62}\scriptsize$\pm$\textbf{0.71}\\
            \underline{{Heartbleed}} & 0.00\scriptsize{$\pm$0.00}& 45.0\scriptsize{$\pm$16.43 }& 73.33\scriptsize{$\pm$1.49 }& 0.00\scriptsize{$\pm$0.00}& 30.0\scriptsize{$\pm$16.43 }& 79.44\scriptsize{$\pm$2.56 }& 0.00\scriptsize{$\pm$0.00}& \cellcolor{LightBlue}\textbf{80.89}\scriptsize$\pm$\textbf{3.64}\\            
            \bottomrule
            \textbf{Average} $\uparrow$ & 67.74\scriptsize{$\pm$0.09 }& 70.59\scriptsize{$\pm$2.63 }& 73.26\scriptsize{$\pm$0.36 }& 67.7\scriptsize{$\pm$0.08 }& 69.66\scriptsize{$\pm$2.49 }& 72.77\scriptsize{$\pm$0.32 }& 67.80\scriptsize{$\pm$0.22 }& \cellcolor{LightBlue}\textbf{74.31}\scriptsize$\pm$\textbf{0.49} \\
            \bottomrule
	\end{tabular}
        }
	\label{table-CIC-IDS2017-F1-adj-0.25}
    \vspace{-0.3cm}
\end{table*}

\setlength{\tabcolsep}{3.0pt}
\begin{table*}[htbp]
	\centering
	\caption{
		\textbf{\textit{F1-score}s obtained by different methods on NSL-KDD dataset (5 runs).} 
        The underlined traffic category belongs to the minority class.
        The results are expressed in \% with standard deviations.}
    \vspace{0.1cm}
    \scalebox{0.892}{
	\begin{tabular}{c|ccccccc|c}
		\toprule
		Traffic Category & {ERM}  & {Focal} & {CB-SGM} & {Mixup-DRW} & {Remix-DRW} & {LDAM-DRW} & {LDR-KL} & {\texttt{\textbf{GDR-CIL}} (Ours)}\\
		\toprule
		Normal & \cellcolor{LightBlue}\textbf{99.27}\scriptsize{$\pm$\textbf{0.02}	}& 99.16\scriptsize{$\pm$0.02 }& 99.23\scriptsize{$\pm$0.01 }& 99.24\scriptsize{$\pm$0.01	}& 99.23\scriptsize{$\pm$0.01 }& \cellcolor{LightBlue}\textbf{99.27}\scriptsize{$\pm$\textbf{0.03} }& 99.08\scriptsize{$\pm$0.02 }& 97.82\scriptsize$\pm$0.03\\
            Probe & \cellcolor{LightBlue}\textbf{96.69}\scriptsize{$\pm$\textbf{0.04}}& 96.42\scriptsize{$\pm$0.04}& 96.56\scriptsize{$\pm$0.04}& 96.66\scriptsize{$\pm$0.05}& 96.60\scriptsize{$\pm$0.02}& 96.47\scriptsize{$\pm$0.30}& 96.63\scriptsize{$\pm$0.03}& 94.76\scriptsize$\pm$0.04\\
            DoS & 97.29\scriptsize{$\pm$0.02 }& 97.26\scriptsize{$\pm$0.05}& 97.32\scriptsize{$\pm$0.04}& 97.37\scriptsize{$\pm$0.07}& 97.26\scriptsize{$\pm$0.07}& \cellcolor{LightBlue}\textbf{97.38}\scriptsize{$\pm$\textbf{0.09}}& 97.23\scriptsize{$\pm$0.04}& 96.34\scriptsize$\pm$0.08\\
            R2L & 90.00\scriptsize{$\pm$0.18 }& 89.53\scriptsize{$\pm$0.11}& 89.43\scriptsize{$\pm$0.12}& 89.84\scriptsize{$\pm$0.13}& 89.83\scriptsize{$\pm$0.11}& \cellcolor{LightBlue}\textbf{90.27}\scriptsize{$\pm$\textbf{0.38}}& 89.85\scriptsize{$\pm$0.04}& 87.89\scriptsize$\pm$0.12\\
            \underline{{U2R}} & 0.00\scriptsize{$\pm$0.00}& 3.41\scriptsize{$\pm$1.87}& 40.07\scriptsize{$\pm$9.03}& 0.00\scriptsize{$\pm$0.00}& 0.00\scriptsize{$\pm$0.00}& 35.66\scriptsize{$\pm$10.34}& 0.00\scriptsize{$\pm$0.00}& \cellcolor{LightBlue}\textbf{49.19}\scriptsize$\pm$\textbf{0.34}\\
            \bottomrule
            \textbf{Average} $\uparrow$ & 76.65\scriptsize{$\pm$0.11}& 77.15\scriptsize{$\pm$0.85}& 84.52\scriptsize{$\pm$1.80}& 76.62\scriptsize{$\pm$0.11}& 	76.58\scriptsize{$\pm$0.08}& 83.81\scriptsize{$\pm$2.16}& 76.56\scriptsize{$\pm$0.03}& \cellcolor{LightBlue}\textbf{85.20}\scriptsize$\pm$\textbf{0.08} \\
            \bottomrule
	\end{tabular}
	}
    \label{table-NSL-KDD-F1}
\end{table*}

\setlength{\tabcolsep}{3.0pt}
\begin{table*}[htbp]
	\centering
	\caption{
		\textbf{\textit{F1-score}s obtained by different methods on UNSW-NB15 dataset (5 runs).}
  Underlined traffic categories belong to the minority class.
  The results are expressed in \% with standard deviations.}
    \vspace{0.1cm}
    \scalebox{0.892}{
	\begin{tabular}{c|ccccccc|c}
		\toprule
		Traffic Category & {ERM}  & {Focal} & {CB-SGM} & {Mixup-DRW} & {Remix-DRW} & {LDAM-DRW} & {LDR-KL} & {\texttt{\textbf{GDR-CIL}} (Ours)}\\
		\toprule
            Normal & 98.53\scriptsize{$\pm$0.02 }& 98.53\scriptsize{$\pm$0.02 }& 98.54\scriptsize{$\pm$0.02 }& 98.54\scriptsize{$\pm$0.01}& 98.52\scriptsize{$\pm$0.01 }& \cellcolor{LightBlue}\textbf{98.57}\scriptsize{$\pm$\textbf{0.00} }& 98.53\scriptsize{$\pm$0.02 }& 98.54\scriptsize$\pm$0.01 \\		
            Generic & 82.87\scriptsize{$\pm$0.05 }& 82.84\scriptsize{$\pm$0.03 }& 82.88\scriptsize{$\pm$0.04 }& \cellcolor{LightBlue}\textbf{82.94}\scriptsize{$\pm$\textbf{0.07} }& \cellcolor{LightBlue}\textbf{82.94}\scriptsize{$\pm$\textbf{0.03} }& 82.74\scriptsize{$\pm$0.02 }& 82.81\scriptsize{$\pm$0.05 }& 82.71\scriptsize$\pm$0.01\\
            Exploits & 77.35\scriptsize{$\pm$0.15 }& 77.69\scriptsize{$\pm$0.27 }& 76.83\scriptsize{$\pm$0.39 }& 76.69\scriptsize{$\pm$0.24 }& 77.01\scriptsize{$\pm$0.21 }& 78.40\scriptsize{$\pm$0.11 }& \cellcolor{LightBlue}\textbf{79.32}\scriptsize{$\pm$\textbf{0.10} }& 77.08\scriptsize$\pm$0.27\\
            Fuzzers & 74.40\scriptsize{$\pm$0.10 }& 74.44\scriptsize{$\pm$0.12 }& 74.45\scriptsize{$\pm$0.12 }& 74.34\scriptsize{$\pm$0.06}& 74.40\scriptsize{$\pm$0.04 }& 74.21\scriptsize{$\pm$0.03 }& \cellcolor{LightBlue}\textbf{74.62}\scriptsize{$\pm$\textbf{0.05} }& 73.82\scriptsize$\pm$0.26\\
            DoS & 58.17\scriptsize{$\pm$0.15 }& 58.05\scriptsize{$\pm$0.08 }& 58.04\scriptsize{$\pm$0.17 }& 57.97\scriptsize{$\pm$0.15 }& 58.08\scriptsize{$\pm$0.05 }& 58.16\scriptsize{$\pm$0.11 }& \cellcolor{LightBlue}\textbf{58.54}\scriptsize{$\pm$\textbf{0.09} }& 58.05\scriptsize$\pm$0.12\\
            Reconnaissance & 56.76\scriptsize{$\pm$0.09 }& 56.87\scriptsize{$\pm$0.16 }& 56.57\scriptsize{$\pm$0.06 }& 56.58\scriptsize{$\pm$0.10 }& 56.48\scriptsize{$\pm$0.04 }& \cellcolor{LightBlue}\textbf{57.25}\scriptsize{$\pm$\textbf{0.14} }& 57.14\scriptsize{$\pm$0.22 }& 55.84\scriptsize$\pm$0.24\\
            \underline{{Analysis}} & 31.83\scriptsize{$\pm$0.15 }& 31.53\scriptsize{$\pm$0.24 }& 32.03\scriptsize{$\pm$0.1 }& 31.73\scriptsize{$\pm$0.17 }& 31.61\scriptsize{$\pm$0.18 }& 32.54\scriptsize{$\pm$0.14 }& \cellcolor{LightBlue}\textbf{32.59}\scriptsize{$\pm$\textbf{0.15} }& 27.77\scriptsize$\pm$0.06\\
            \underline{{Backdoor}} & 0.00\scriptsize{$\pm$0.00}& 1.37\scriptsize{$\pm$0.76 }& 0.00\scriptsize{$\pm$0.00}&  0.00\scriptsize{$\pm$0.00}&  0.00\scriptsize{$\pm$0.00}& 0.06\scriptsize{$\pm$0.05 }&  0.00\scriptsize{$\pm$0.00}& \cellcolor{LightBlue}\textbf{23.17}\scriptsize$\pm$\textbf{0.28}\\
            \underline{{Shellcode}} & 38.54\scriptsize{$\pm$0.21 }& 41.64\scriptsize{$\pm$1.15 }& 38.02\scriptsize{$\pm$0.46 }& 37.71\scriptsize{$\pm$0.42 }& 38.46\scriptsize{$\pm$0.57 }& 43.76\scriptsize{$\pm$1.25 }& \cellcolor{LightBlue}\textbf{45.82}\scriptsize{$\pm$\textbf{0.93} }& 45.23\scriptsize$\pm$0.94\\
            \underline{{Worms}} & 5.28\scriptsize{$\pm$4.73 }& 15.67\scriptsize{$\pm$5.73 }& 19.79\scriptsize{$\pm$4.46 }& 6.15\scriptsize{$\pm$4.59	 }& 10.38\scriptsize{$\pm$5.69 }& \cellcolor{LightBlue}\textbf{22.51}\scriptsize{$\pm$\textbf{1.42} }& 0.00\scriptsize{$\pm$0.00}& 21.61\scriptsize$\pm$0.57\\
            \bottomrule
            \textbf{Average} $\uparrow$ & 52.37\scriptsize{$\pm$1.17 }& 53.86\scriptsize{$\pm$0.68 }& 53.71\scriptsize{$\pm$1.11 }& 52.26\scriptsize{$\pm$1.16	 }& 52.79\scriptsize{$\pm$1.42 }& 54.82\scriptsize{$\pm$0.26 }& 52.94\scriptsize{$\pm$0.30 }& \cellcolor{LightBlue}\textbf{56.38}\scriptsize$\pm$\textbf{0.23}\\
            \bottomrule
	\end{tabular}
        }
	\label{table-UNSW-NB15-F1}
\end{table*}

\setlength{\tabcolsep}{3.0pt}
\begin{table*}[htbp]
	\centering
	\caption{
		\textbf{\textit{G-mean}s obtained by different methods on CIC-IDS2017 dataset (5 runs).} 
  Underlined traffic categories belong to the minority class.
  The results are expressed in \% with standard deviations.}
    \vspace{0.1cm}
    \scalebox{0.84}{
	\begin{tabular}{c|ccccccc|c}
		\toprule
		Traffic Category & {ERM}  & {Focal} & {CB-SGM} & {Mixup-DRW} & {Remix-DRW} & {LDAM-DRW} & {LDR-KL} & {\texttt{\textbf{GDR-CIL}} (Ours)}\\
		\toprule
		BENIGN        & \cellcolor{LightBlue}\textbf{96.88}\scriptsize{$\pm$\textbf{0.07}	}&96.67\scriptsize{$\pm$0.04	}&96.62\scriptsize{$\pm$0.05	}&96.86\scriptsize{$\pm$0.07}&	96.86\scriptsize{$\pm$0.05}&	96.57\scriptsize{$\pm$0.29}&	96.64\scriptsize{$\pm$0.17	}& 94.83\scriptsize$\pm$0.06\\
            DoS Hulk      & 96.44\scriptsize{$\pm$0.40 }&	95.71\scriptsize{$\pm$0.23 }&	95.99\scriptsize{$\pm$0.32}&	96.38\scriptsize{$\pm$0.42}&	96.63\scriptsize{$\pm$0.39}&	\cellcolor{LightBlue}\textbf{96.93}\scriptsize{$\pm$\textbf{0.05}}&	96.6\scriptsize{$\pm$0.28 }&	96.18\scriptsize$\pm$0.03\\
            DDoS          & 98.82\scriptsize{$\pm$0.03 }&	98.54\scriptsize{$\pm$0.05 }&	98.41\scriptsize{$\pm$0.03}&	98.84\scriptsize{$\pm$0.09}&	98.81\scriptsize{$\pm$0.02}&	98.84\scriptsize{$\pm$0.07}&	\cellcolor{LightBlue}\textbf{99.19}\scriptsize{$\pm$\textbf{0.02} }&	97.00\scriptsize$\pm$0.07\\
            PortScan      & \cellcolor{LightBlue}\textbf{99.24}\scriptsize{$\pm$\textbf{0.00}}&	99.22\scriptsize{$\pm$0.01}&	99.03\scriptsize{$\pm$0.05}&	99.23\scriptsize{$\pm$0.00}&	99.24\scriptsize{$\pm$0.00}&	\cellcolor{LightBlue}\textbf{99.24}\scriptsize{$\pm$\textbf{0.00}}&	97.12\scriptsize{$\pm$0.91}&	98.81\scriptsize$\pm$0.10\\
            \underline{{DoS GoldenEye}} & 99.26\scriptsize{$\pm$0.04 }&	99.21\scriptsize{$\pm$0.00}& 99.18\scriptsize{$\pm$0.02}&	99.25\scriptsize{$\pm$0.03}&	99.21\scriptsize{$\pm$0.00}&	99.23\scriptsize{$\pm$0.20}&	\cellcolor{LightBlue}\textbf{99.47}\scriptsize{$\pm$\textbf{0.00} }&	98.68\scriptsize$\pm$0.02\\
            \underline{{FTP-Patator}}   & 99.65\scriptsize{$\pm$0.00}&	\cellcolor{LightBlue}\textbf{99.67}\scriptsize{$\pm$\textbf{0.00}}&	99.64\scriptsize{$\pm$0.01}&	99.65\scriptsize{$\pm$0.00}&	99.65\scriptsize{$\pm$0.01}&	99.58\scriptsize{$\pm$0.02}&	99.64\scriptsize{$\pm$0.02}&	99.67\scriptsize$\pm$0.01\\
		\underline{DoS slowloris} & 97.85\scriptsize{$\pm$0.00}&	\cellcolor{LightBlue}\textbf{98.13}\scriptsize{$\pm$\textbf{0.14}}&	97.29\scriptsize{$\pm$0.14}&	97.86\scriptsize{$\pm$0.01}&	97.88\scriptsize{$\pm$0.02}&	97.89\scriptsize{$\pm$0.03}&	97.94\scriptsize{$\pm$0.11}&	97.85\scriptsize$\pm$0.04\\
            \underline{{DoS Slowhttptest}} & 99.19\scriptsize{$\pm$0.01}&	\cellcolor{LightBlue}\textbf{99.22}\scriptsize{$\pm$\textbf{0.01}}&	98.91\scriptsize{$\pm$0.01}&	99.18\scriptsize{$\pm$0.02}&	99.17\scriptsize{$\pm$0.02}&	99.14\scriptsize{$\pm$0.03}&	99.20\scriptsize{$\pm$0.01}&	99.14\scriptsize$\pm$0.02\\
            \underline{{SSH-Patator}}      & 95.63\scriptsize{$\pm$0.00}& 95.63\scriptsize{$\pm$0.00}& 95.62\scriptsize{$\pm$0.00}& 95.64\scriptsize{$\pm$0.01 }&	95.64\scriptsize{$\pm$0.01 }&	\cellcolor{LightBlue}\textbf{95.65}\scriptsize{$\pm$\textbf{0.01} }&	95.64\scriptsize{$\pm$0.00}&	\cellcolor{LightBlue}\textbf{95.65}\scriptsize$\pm$\textbf{0.00}\\
            \underline{{Bot}}           & 81.62\scriptsize{$\pm$0.03 }&	80.25\scriptsize{$\pm$0.71 }&	78.15\scriptsize{$\pm$0.04 }&	81.58\scriptsize{$\pm$0.02 }&	81.41\scriptsize{$\pm$0.17 }&	84.87\scriptsize{$\pm$2.98 }&	81.11\scriptsize{$\pm$0.24 }&	\cellcolor{LightBlue}\textbf{98.91}\scriptsize$\pm$\textbf{0.04}\\
           \underline{{Web Attack Brute Force}} & 92.31\scriptsize{$\pm$0.00}&	93.15\scriptsize{$\pm$0.50 }&	92.04\scriptsize{$\pm$0.11 }&	92.32\scriptsize{$\pm$0.00}&	92.32\scriptsize{$\pm$0.00}&	81.04\scriptsize{$\pm$10.98 }&	\cellcolor{LightBlue}\textbf{93.41}\scriptsize{$\pm$\textbf{0.42} }&	76.45\scriptsize$\pm$3.51\\
            \underline{{Web Attack XSS}}   & 0.00\scriptsize{$\pm$0.00}&	0.00\scriptsize{$\pm$0.00}&	0.00\scriptsize{$\pm$0.00}&	0.00\scriptsize{$\pm$0.00}&	0.00\scriptsize{$\pm$0.00}&	22.20\scriptsize{$\pm$16.63}&	0.00\scriptsize{$\pm$0.00}&	\cellcolor{LightBlue}\textbf{77.34}\scriptsize$\pm$\textbf{6.87}\\
		\underline{{Infiltration}}     & 0.00\scriptsize{$\pm$0.00}&	0.00\scriptsize{$\pm$0.00}&	28.02\scriptsize{$\pm$6.55 }&	0.00\scriptsize{$\pm$0.00}&	0.00\scriptsize{$\pm$0.00}&	17.69\scriptsize{$\pm$15.82 }&	0.00\scriptsize{$\pm$0.00}&	\cellcolor{LightBlue}\textbf{72.65}\scriptsize$\pm$\textbf{1.07}\\
            \underline{{Web Attack Sql Injection}} & 0.00\scriptsize{$\pm$0.00}&	0.00\scriptsize{$\pm$0.00}&	0.00\scriptsize{$\pm$0.00}&	0.00\scriptsize{$\pm$0.00}&	0.00\scriptsize{$\pm$0.00}&	8.93\scriptsize{$\pm$7.98 }&	0.00\scriptsize{$\pm$0.00}&	\cellcolor{LightBlue}\textbf{59.51}\scriptsize$\pm$\textbf{3.31}\\
            \underline{{Heartbleed}}       & 0.00\scriptsize{$\pm$0.00}&	46.48\scriptsize{$\pm$16.97 }&	77.46\scriptsize{$\pm$0.00}&	0.00\scriptsize{$\pm$0.00}&	30.98\scriptsize{$\pm$16.97 }&	84.36\scriptsize{$\pm$4.07 }&	0.00\scriptsize{$\pm$0.00}&	\cellcolor{LightBlue}\textbf{89.44}\scriptsize$\pm$\textbf{0.00}\\            
            \bottomrule
            \textbf{Average} $\uparrow$       & 70.46\scriptsize{$\pm$0.03 }&	73.46\scriptsize{$\pm$1.21 }&	77.09\scriptsize{$\pm$0.42 }&	70.45\scriptsize{$\pm$0.04 }&	72.52\scriptsize{$\pm$1.15 }&	78.81\scriptsize{$\pm$2.35 }&	70.4\scriptsize{$\pm$0.09 }&	\cellcolor{LightBlue}\textbf{90.14}\scriptsize$\pm$\textbf{0.43} \\
            \bottomrule
	\end{tabular}
        }
	\label{table-CIC-IDS2017-Gmean-adj-0.25}
    \vspace{-0.3cm}
\end{table*}

\setlength{\tabcolsep}{3.0pt}
\begin{table*}[htbp]
	\centering
	\caption{
		\textbf{\textit{G-mean}s obtained by different methods on NSL-KDD dataset (5 runs).} 
        The underlined traffic category belongs to the minority class.
        The results are expressed in \% with standard deviations.}
    \vspace{0.1cm}
    \scalebox{0.892}{
	\begin{tabular}{c|ccccccc|c}
		\toprule
		Traffic Category & {ERM}  & {Focal} & {CB-SGM} & {Mixup-DRW} & {Remix-DRW} & {LDAM-DRW} & {LDR-KL} & {\texttt{\textbf{GDR-CIL}} (Ours)}\\
		\toprule
            Normal &99.38\scriptsize{$\pm$0.01 }&	99.32\scriptsize{$\pm$0.01 }&	99.35\scriptsize{$\pm$0.01 }&	99.37\scriptsize{$\pm$0.01 }&	99.36\scriptsize{$\pm$0.01 }&	\cellcolor{LightBlue}\textbf{99.41}\scriptsize{$\pm$\textbf{0.03} }&	99.26\scriptsize{$\pm$0.02 }&	97.97\scriptsize$\pm$0.02\\
            Probe &\cellcolor{LightBlue}\textbf{97.08}\scriptsize{$\pm$\textbf{0.04} }&	96.80\scriptsize{$\pm$0.04 }&	96.96\scriptsize{$\pm$0.04 }&	97.06\scriptsize{$\pm$0.05 }&	97.01\scriptsize{$\pm$0.02 }&	96.79\scriptsize{$\pm$0.31 }&	97.03\scriptsize{$\pm$0.04 }&	95.37\scriptsize$\pm$0.04\\
            DoS &98.92\scriptsize{$\pm$0.01 }&	\cellcolor{LightBlue}\textbf{99.00}\scriptsize{$\pm$\textbf{0.02} }&	98.91\scriptsize{$\pm$0.02 }&	98.94\scriptsize{$\pm$0.01 }&	98.90\scriptsize{$\pm$0.02 }&	\cellcolor{LightBlue}\textbf{99.00}\scriptsize{$\pm$\textbf{0.05} }&	98.92\scriptsize{$\pm$0.02 }&	98.50\scriptsize$\pm$0.07\\
            R2L &95.63\scriptsize{$\pm$0.06 }&	95.97\scriptsize{$\pm$0.07 }&	94.98\scriptsize{$\pm$0.10 }&	95.38\scriptsize{$\pm$0.09 }&	95.43\scriptsize{$\pm$0.06 }&	96.29\scriptsize{$\pm$0.07 }&	95.22\scriptsize{$\pm$0.18 }&	\cellcolor{LightBlue}\textbf{96.94}\scriptsize$\pm$\textbf{0.07}\\
            \underline{U2R} &0.00\scriptsize{$\pm$0.00 }&	8.53\scriptsize{$\pm$4.67 }&	56.51\scriptsize{$\pm$12.67 }&	0.00\scriptsize{$\pm$0.00}&	0.00\scriptsize{$\pm$0.00 }&	60.48\scriptsize{$\pm$13.80 }&	0.00\scriptsize{$\pm$0.00 }&	\cellcolor{LightBlue}\textbf{73.81}\scriptsize$\pm$\textbf{0.00}\\
            \bottomrule
            \textbf{Average} $\uparrow$ & 78.20\scriptsize{$\pm$0.02 }&	79.92\scriptsize{$\pm$0.94 }&	89.34\scriptsize{$\pm$2.53 }&	78.15\scriptsize{$\pm$0.02 }&	78.14\scriptsize{$\pm$0.01 }&	90.40\scriptsize{$\pm$2.75 }&	78.08\scriptsize{$\pm$0.03 }&	\cellcolor{LightBlue}\textbf{92.52}\scriptsize$\pm$\textbf{0.02} \\
            \bottomrule
	\end{tabular}
        }
	\label{table-NSL-KDD-Gmean}
    \vspace{-0.3cm}
\end{table*}

\setlength{\tabcolsep}{3.0pt}
\begin{table*}[htbp]
	\centering
	\caption{
		\textbf{\textit{G-mean}s obtained by different methods on UNSW-NB15 dataset (5 runs).}
  Underlined traffic categories belong to the minority class.
  The results are expressed in \% with standard deviations.}
    \vspace{0.1cm}
    \scalebox{0.892}{
	\begin{tabular}{c|ccccccc|c}
		\toprule
		Traffic Category & {ERM}  & {Focal} & {CB-SGM} & {Mixup-DRW} & {Remix-DRW} & {LDAM-DRW} & {LDR-KL} & {\texttt{\textbf{GDR-CIL}} (Ours)}\\
		\toprule
            Normal & \cellcolor{LightBlue}\textbf{98.71}\scriptsize{$\pm$\textbf{0.00} }& 98.70\scriptsize{$\pm$0.00}& 	98.7\scriptsize{$\pm$0.00}& 	98.70\scriptsize{$\pm$0.00}& 	98.7\scriptsize{$\pm$0.00}& \cellcolor{LightBlue}\textbf{98.71}\scriptsize{$\pm$\textbf{0.00}	 }& \cellcolor{LightBlue}\textbf{98.71}\scriptsize{$\pm$\textbf{0.00}	 }& 98.69\scriptsize$\pm$0.00\\		
            Generic & 84.34\scriptsize{$\pm$0.04 }& 	84.30\scriptsize{$\pm$0.04 }& 	84.37\scriptsize{$\pm$0.07 }& 	\cellcolor{LightBlue}\textbf{84.41}\scriptsize{$\pm$\textbf{0.09} }& 	84.39\scriptsize{$\pm$0.04 }& 	84.07\scriptsize{$\pm$0.01 }& 	84.16\scriptsize{$\pm$0.09 }& 	83.97\scriptsize$\pm$0.01\\
            Exploits & 86.79\scriptsize{$\pm$0.08	 }& 86.95\scriptsize{$\pm$0.13 }& 	86.59\scriptsize{$\pm$0.11 }& 	86.68\scriptsize{$\pm$0.03	 }& 86.80\scriptsize{$\pm$0.06 }& 	87.06\scriptsize{$\pm$0.06 }& 	\cellcolor{LightBlue}\textbf{87.38}\scriptsize{$\pm$\textbf{0.04} }& 	86.10\scriptsize$\pm$0.22\\
            Fuzzers & 87.87\scriptsize{$\pm$0.07 }& 87.74\scriptsize{$\pm$0.09 }& 87.82\scriptsize{$\pm$0.07 }& 87.71\scriptsize{$\pm$0.07 }& 87.75\scriptsize{$\pm$0.07 }& 87.65\scriptsize{$\pm$0.04 }& \cellcolor{LightBlue}\textbf{87.91}\scriptsize{$\pm$\textbf{0.04} }& 85.89\scriptsize$\pm$0.21\\
            DoS & 68.73\scriptsize{$\pm$0.12 }& 68.71\scriptsize{$\pm$0.14 }& 68.78\scriptsize{$\pm$0.16 }& 68.79\scriptsize{$\pm$0.11 }& 68.88\scriptsize{$\pm$0.07 }& 68.52\scriptsize{$\pm$0.17 }& \cellcolor{LightBlue}\textbf{68.95}\scriptsize{$\pm$\textbf{0.18} }& 68.47\scriptsize$\pm$0.17\\
            Reconnaissance & 80.03\scriptsize{$\pm$0.08 }& 80.13\scriptsize{$\pm$0.16 }& 79.80\scriptsize{$\pm$0.05 }& 79.76\scriptsize{$\pm$0.09	 }& 79.74\scriptsize{$\pm$0.03 }& 80.61\scriptsize{$\pm$0.14 }& \cellcolor{LightBlue}\textbf{80.66}\scriptsize{$\pm$\textbf{0.29} }& 77.90\scriptsize$\pm$0.38\\
            \underline{Analysis} & 52.85\scriptsize{$\pm$0.29 }& 52.49\scriptsize{$\pm$0.33 }& 53.35\scriptsize{$\pm$0.18 }& 52.84\scriptsize{$\pm$0.30	 }& 52.45\scriptsize{$\pm$0.33 }& 54.33\scriptsize{$\pm$0.24 }& \cellcolor{LightBlue}\textbf{54.51}\scriptsize{$\pm$\textbf{0.20} }& 44.82\scriptsize$\pm$0.10\\
            \underline{Backdoor} & 0.00\scriptsize{$\pm$0.00}& 5.44\scriptsize{$\pm$3.00 }& 0.00\scriptsize{$\pm$0.00}&  0.00\scriptsize{$\pm$0.00}&  0.00\scriptsize{$\pm$0.00}& 0.74\scriptsize{$\pm$0.67 }&  0.00\scriptsize{$\pm$0.00}& \cellcolor{LightBlue}\textbf{50.86}\scriptsize$\pm$\textbf{0.53}\\
            \underline{Shellcode} & 54.47\scriptsize{$\pm$0.70 }& 56.95\scriptsize{$\pm$0.76 }& 52.82\scriptsize{$\pm$0.95 }& 52.47\scriptsize{$\pm$0.81 }& 53.42\scriptsize{$\pm$0.90 }& 59.66\scriptsize{$\pm$1.09 }& 59.36\scriptsize{$\pm$0.83 }& \cellcolor{LightBlue}\textbf{67.92}\scriptsize$\pm$\textbf{0.73}\\
            \underline{Worms} & 7.89\scriptsize{$\pm$7.06 }& 	23.66\scriptsize{$\pm$8.64 }& 	31.55\scriptsize{$\pm$7.05 }& 	10.87\scriptsize{$\pm$6.89	 }& 15.78\scriptsize{$\pm$8.64 }& 	37.51\scriptsize{$\pm$1.72	 }& 0.00\scriptsize{$\pm$0.00}& \cellcolor{LightBlue}\textbf{39.43}\scriptsize$\pm$\textbf{0.00}\\
            \bottomrule
            \textbf{Average} $\uparrow$ & 62.17\scriptsize{$\pm$0.78	 }& 64.51\scriptsize{$\pm$1.12 }& 	64.38\scriptsize{$\pm$0.78 }& 	62.22\scriptsize{$\pm$0.74	 }& 62.79\scriptsize{$\pm$0.94 }& 	65.89\scriptsize{$\pm$0.31 }& 	62.16\scriptsize{$\pm$0.11 }& 	\cellcolor{LightBlue}\textbf{70.41}\scriptsize$\pm$\textbf{0.09}\\
            \bottomrule
	\end{tabular}
        }
	\label{table-UNSW-NB15-Gmean}
    \vspace{-0.3cm}
\end{table*}

In the field of imbalance network traffic classification, comprehensive indicators are commonly used in performance evaluation.
Hence, we respectively report comprehensive indicators, such as \textit{Specificity}, \textit{F1-score}, \textit{G-mean}, \textit{AUC} on all benchmarks' testing datasets using various classification models.

\textbf{Result analysis from \textit{Specificity}.}
Fig. \ref{Specificity-bar} indicates the \textit{Specificity} comparisons between \texttt{\textbf{GDR-CIL}} and the other seven baselines.
\texttt{\textbf{GDR-CIL}} surpasses all other baselines on the two extremely imbalanced datasets, CIC-IDS2017 and UNSW-NB15.
For the NSL-KDD dataset, a less imbalanced dataset with only five classes, the grouping mechanism in \texttt{\textbf{GDR-CIL}} encounters limitations in improving performance.
As a result, \texttt{\textbf{GDR-CIL}} exhibits a slight degradation in the average \textit{Specificity}.

\textbf{Result analysis from \textit{F1-score}.}
As observed in Table \ref{table-CIC-IDS2017-F1-adj-0.25}, all baselines except ours achieve relatively high average performance on \textit{F1-score} for all majority and some minority classes on CIC-IDS2017 dataset. 
However, most of them yield low \textit{F1-score}s in certain minority classes, such as ``Web Attack XSS'' class where the \textit{F1-score} is almost 0. 
In comparison, \texttt{\textbf{GDR-CIL}}'s performance is comparable to other baseline methods on major classes while retaining discriminative capability for the minority classes.
Particularly, in minor but more risky classes like ``Web Attack XSS'', ``Infiltration'' and ``Web Attack Sql Injection'', our approach exhibits a significant improvement over several baselines which achieve \textit{F1-score}s nearly 0, e.g., notably increasing the score by at least $31.87\%$ in ``Web Attack XSS'' class over LDAW-DRW.

Additionally, it can be observed that there is significant variability in \textit{F1-score}s for certain traffic categories, particularly for minority classes.
Some minority classes, like ``SSH-Patator'', display more distinctive traffic patterns, and all methods demonstrate effective discrimination for them.
However, other minority classes with very few instances, such as ``Web Attack Brute Force'', may partially overlap in characteristics with normal traffic or other attack traffic.
This overlap makes it challenging for all methods to differentiate between them, leading to a considerable decline in \textit{F1-score}s.

Note that LDAM-DRW exhibits a subtle ability to recognize all minority classes benefiting from class-dependent margin constraint, while \texttt{\textbf{GDR-CIL}} enhances the recognition performance for hard-to-classify minority classes by reweighting losses at the group level.
Additionally, \texttt{\textbf{GDR-CIL}} outperforms all baselines in average \textit{F1-score}, illustrating its strength in mitigating the negative effect of class imbalance.
The above evidence underscores the necessity of employing group distributionally robust optimization in addressing network traffic classification with numerous minority classes.

For NSL-KDD and UNSW-NB15 datasets with fewer minority classes, we report the \textit{F1-score}s in Tables \ref{table-NSL-KDD-F1}/\ref{table-UNSW-NB15-F1}.
On the NSL-KDD dataset, the baselines ERM, Mixup-DRW, Remix-DRW, and LDR-KL exhibit no identification capability for the minority class ``U2R'', resulting in $ \textit{F1-score} = 0$. 
The remaining baselines cannot achieve \textit{F1-score}s higher than $41\%$.
In contrast, \texttt{\textbf{GDR-CIL}} still obtains nearly $49.19\%$ \textit{F1-score} on the minority class ``U2R'' and simultaneously retains comparable performance to other baselines on four majority classes.
On the UNSW-NB15 dataset, \texttt{\textbf{GDR-CIL}} beats all baselines on average $ \textit{F1-score}$, likely attributed to its remarkable performance gains in the minority class ``Backdoor''.

\textbf{Result analysis from \textit{G-mean}.}
Tables \ref{table-CIC-IDS2017-Gmean-adj-0.25}-\ref{table-UNSW-NB15-Gmean} illustrate the phenomenon similar to that of collected results in \textit{F1-score} in statistics.
Without sacrificing too much identification of the majority classes, \texttt{\textbf{GDR-CIL}} consistently outperforms all other baselines in terms of average \textit{G-mean} for most minority classes.
This further examines the significance of group distributionally robust optimization in suppressing the negative influence of class imbalance. 

\begin{figure*}[t!]
\centerline{\includegraphics[width=1.0\textwidth]{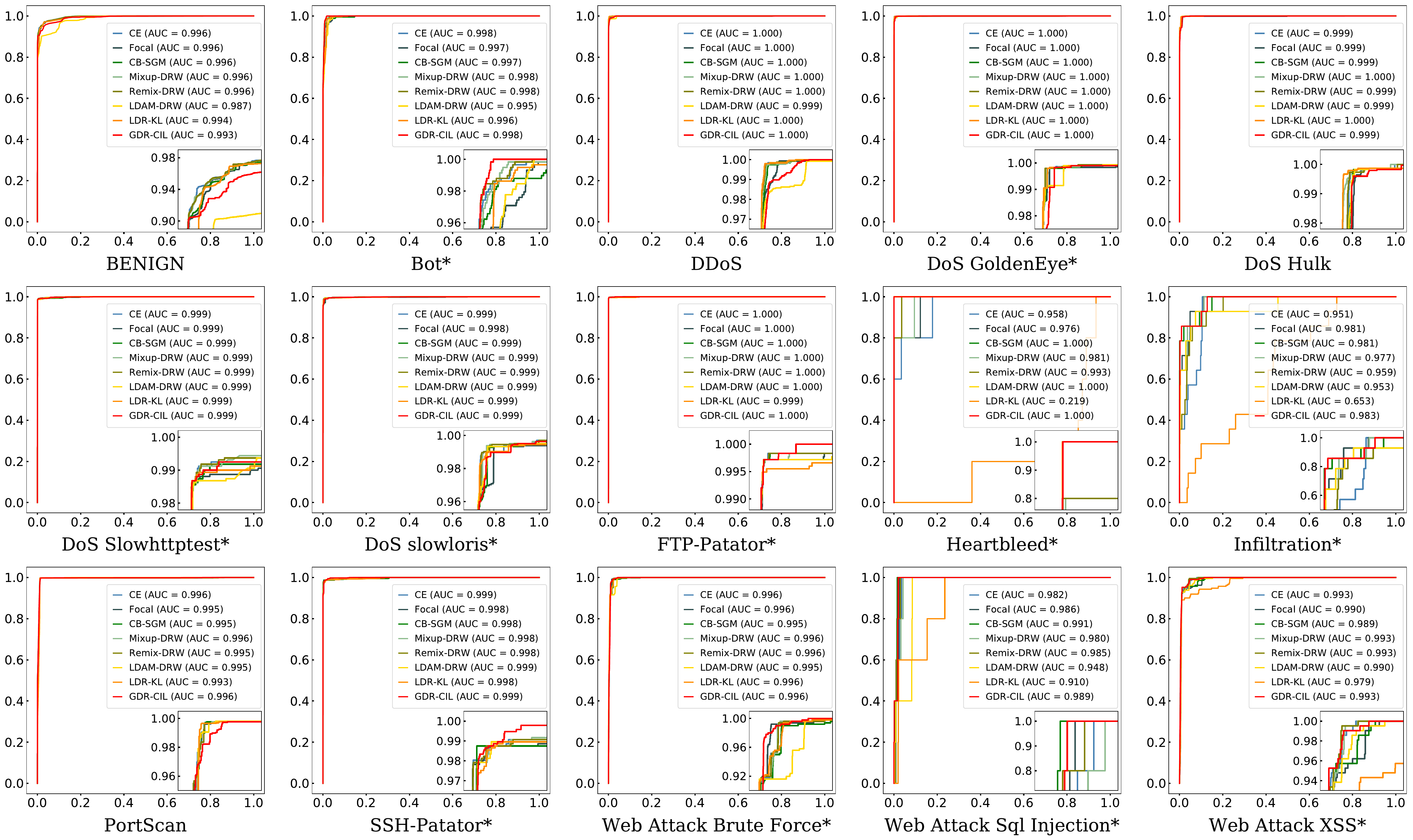}}
\caption{\textbf{ROC curves for each class using different methods on the CIC-IDS2017 dataset.}
Categories marked with an asterisk \textasteriskcentered\ belong to the minority category.}
\label{CIC-IDS2017_ROC-curve}
\end{figure*}

\begin{figure*}[h!]
\centerline{\includegraphics[width=1.0\textwidth]{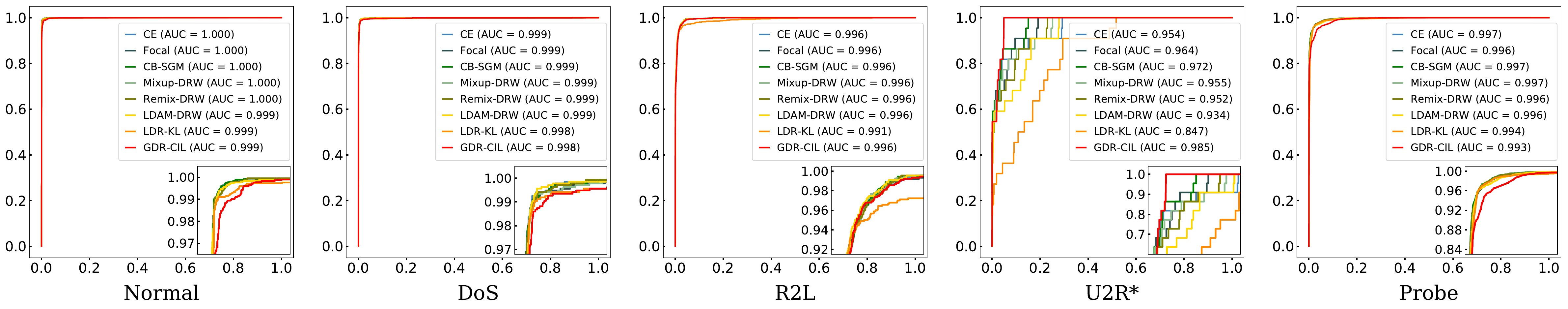}}
\caption{\textbf{ROC curves for each class using different methods on NSL-KDD dataset.}
Categories marked with an asterisk \textasteriskcentered\ belong to the minority category.}
\label{NSL-KDD_ROC-curve}
\end{figure*}

\begin{figure*}[h!]
\centerline{\includegraphics[width=1.0\textwidth]{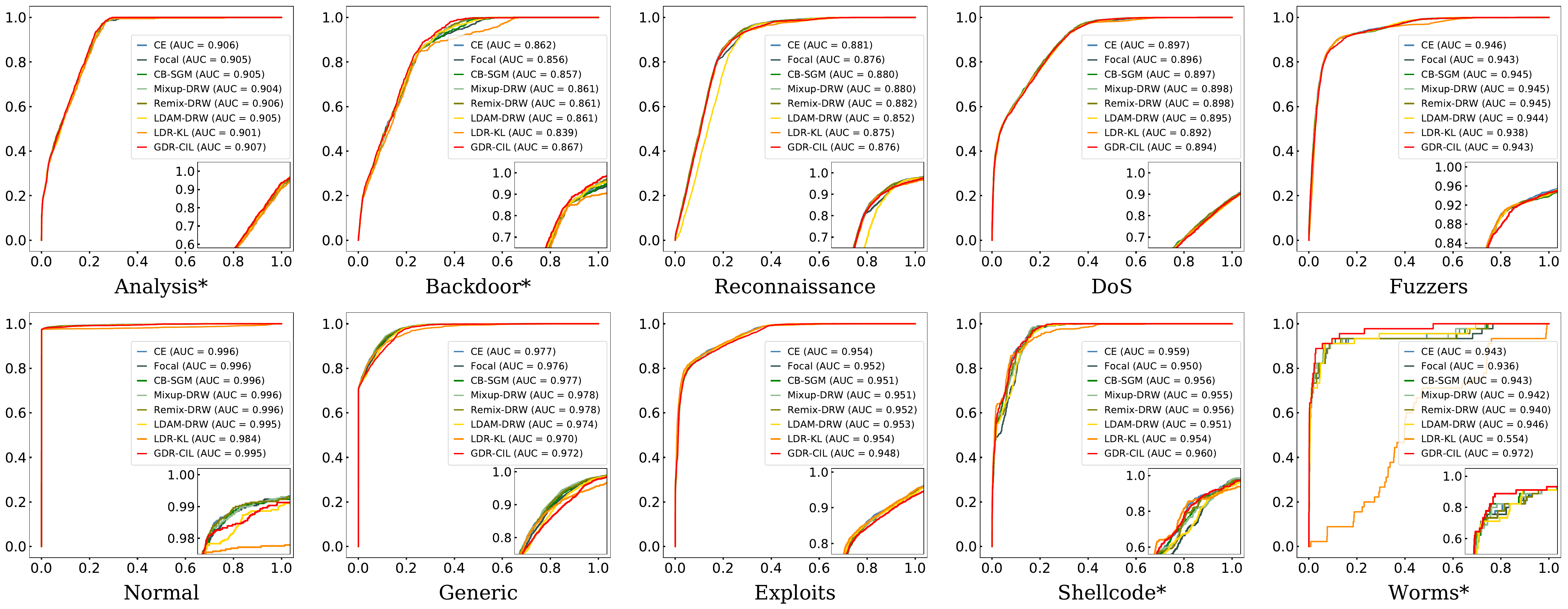}}
\caption{\textbf{ROC curves for each class using different methods on UNSW-NB15 dataset.}
Categories marked with an asterisk \textasteriskcentered\ belong to the minority category.}
\label{UNSW_NB15_ROC-curve}
\end{figure*}

\textbf{Result analysis from \textit{AUC}.}
Fig.s \ref{CIC-IDS2017_ROC-curve}-\ref{UNSW_NB15_ROC-curve} show the ROC curves and average \textit{AUC} values for each class using various methods on the three datasets.
As shown in Fig. \ref{CIC-IDS2017_ROC-curve}, both \texttt{\textbf{GDR-CIL}} and CB-SGM demonstrate high \textit{AUC} values across all classes, including both majority and minority classes.
In contrast, other baseline methods perform poorly on some minority classes, particularly LDR-KL on ``Heartbleed'' and ``Infiltration'' classes.
In comparison, \texttt{\textbf{GDR-CIL}} consistently outperforms all baseline methods, except both on ``Web Attack SQL Injection'' class, where it slightly underperforms CB-SGM with a performance gap of approximately $0.002$, and on ``BENIGN'' class, where it falls short against some methods by a small margin of about $0.003$.
In Fig. \ref{NSL-KDD_ROC-curve}, both \texttt{\textbf{GDR-CIL}} and LDAM-DRW demonstrate higher \textit{AUC} values across all five classes on NSL-KDD dataset, while other methods are less effective on the minority class ``U2R''.
Fig. \ref{UNSW_NB15_ROC-curve} exhibits similar results that all methods yield close \textit{AUC} values, with only minor variations, except for LDR-KL, which performs considerably worse on the minority `` Worms'' class.
It is noteworthy that \texttt{\textbf{GDR-CIL}} demonstrates better performance on minority classes, particularly in the ``Worms'' class, where the other seven baseline methods record lower \textit{AUC} values, with a reduction of over $0.026$.

Overall, the analysis of the above results confirms that our method performs better on the minority classes while maintaining a decent identification capability on the majority ones, especially for datasets with many classes.
This further emphasizes the superiority of group distributionally robust optimization in mitigating the adverse effects of class imbalance.

\subsection{Ablation Studies}\label{Ablation Studies}

Our ablation studies are to examine:
(1) The extent of performance improvement achieved by using the grouping mechanism as compared to not using it;
(2) The feasibility of the heuristic estimate of the number of groups;
(3) The impact of the hyper-parameters in \texttt{\textbf{GDR-CIL}} on the multi-classification performance.
Without loss of generality, we conduct these experiments on CIC-IDS2017 dataset.

\textbf{Effectiveness of the Grouping Mechanism.}
To better understand the role of class clusters, we perform an ablation study by removing the grouping operation in the loss function.
The resulting non-grouped \texttt{\textbf{GDR-CIL}} treats each class as a separate group.
We also take an interest in the calibration $\mathbf{v}_{\mathcal{G}}$ with a hyper-parameter $B$.
Table \ref{table-ablation} compares the grouped \texttt{\textbf{GDR-CIL}} and the non-grouped \texttt{\textbf{GDR-CIL}} in three comprehensive indicators.
We can observe that the two grouped \texttt{\textbf{GDR-CIL}}s with $B=0$ and $B=0.25$ are overall superior to the non-grouped one.
This suggests that the grouping mechanism tends to focus more on the minority classes and correct the decision bias in the imbalanced case.

\setlength{\tabcolsep}{3.0pt}
\begin{table}[htbp]
	\centering
	\caption{
		\textbf{Influence of grouping in \texttt{\textbf{GDR-CIL}} on CIC-IDS2017 dataset (5 runs).} The results are expressed in \% with standard deviations.}
    \vspace{0.1cm}
    \scalebox{0.95}{
	\begin{tabular}{c|ccc}
		\toprule
		   & \textit{F1-score} $\uparrow$  & \textit{G-mean} $\uparrow$ & \textit{Specificity} $\uparrow$ \\
		\toprule
            Non-grouped &70.41\scriptsize{$\pm$0.35 }&	83.01\scriptsize{$\pm$0.17 }&	99.39\scriptsize{$\pm$0.02 }\\
            $K = 8$, $B=0$&72.55\scriptsize{$\pm$0.37 }&	84.02\scriptsize{$\pm$0.65 }&	\cellcolor{LightBlue}\textbf{99.44}\scriptsize{$\pm$\textbf{0.01}}\\
            $K = 8$, $B=0.25$ &\cellcolor{LightBlue}{\textbf{74.31}\scriptsize{$\pm$\textbf{0.49}}}&	\cellcolor{LightBlue}{\textbf{90.14}\scriptsize{$\pm$\textbf{0.43} }}&	\cellcolor{LightBlue}{\textbf{99.44}\scriptsize{$\pm$\textbf{0.01}} }\\
            \bottomrule
	\end{tabular}
        }
	\label{table-ablation}
\end{table}

\begin{figure}[htbp]
\centerline{\includegraphics[width=0.5\textwidth]{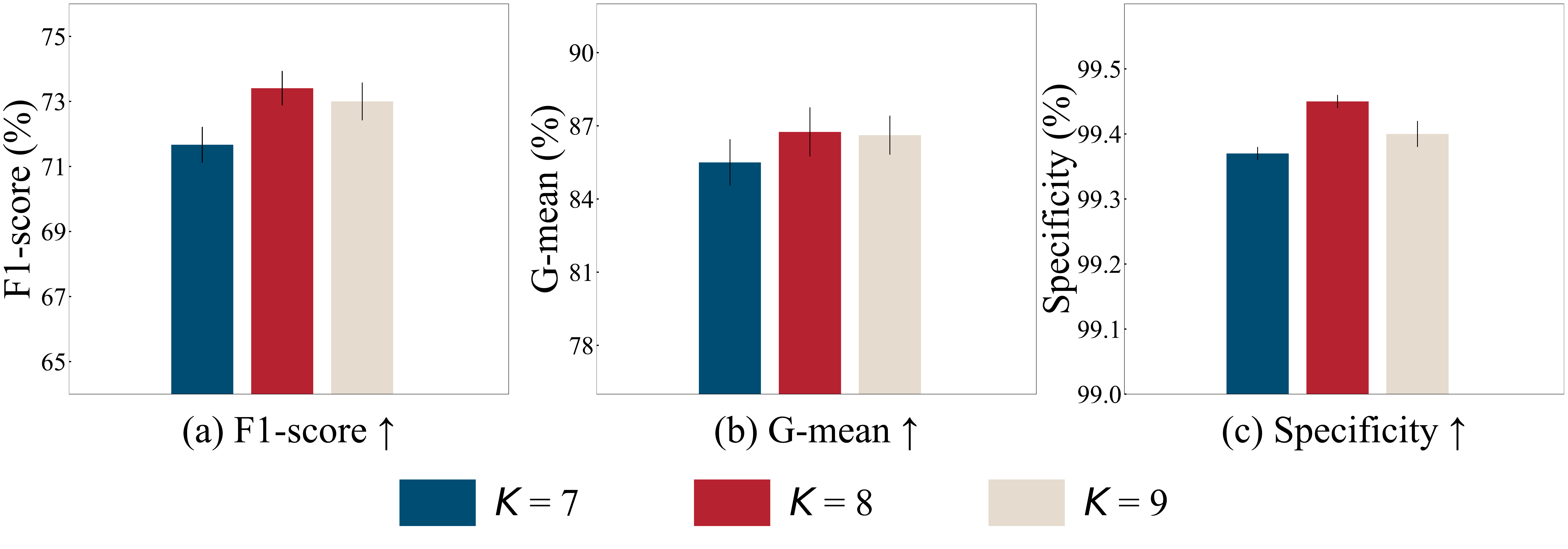}}
\caption{\textbf{The comprehensive performance of \texttt{\textbf{GDR-CIL}} with error bars by varying the number of groups on CIC-IDS2017 dataset (5 runs).}}
\label{CIC-IDS2017_different-groups-f1score-bar}
\end{figure}

\textbf{Feasibility of Heuristic Group Number Estimates.}
We conduct experiments by varying the number of groups $K$ to assess its impact on performance. 
Since different cluster numbers in the K-means algorithm lead to varying group numbers, we evaluate the classification performance under different group numbers by adjusting the number of clusters.
Specifically, $K=7,8$ and $9$ are investigated.
Fig. \ref{CIC-IDS2017_different-groups-f1score-bar} shows the comparison results of \texttt{\textbf{GDR-CIL}}, in which we empirically set $B=0.25$ and $\beta=0.005$.
It reveals the necessity of appropriate selection of the group number, but nearly all cases still reserve superiority over other baseline results in Table \ref{table-CIC-IDS2017-F1-adj-0.25}.

\begin{figure}[tbp]
\centerline{\includegraphics[width=0.5\textwidth]{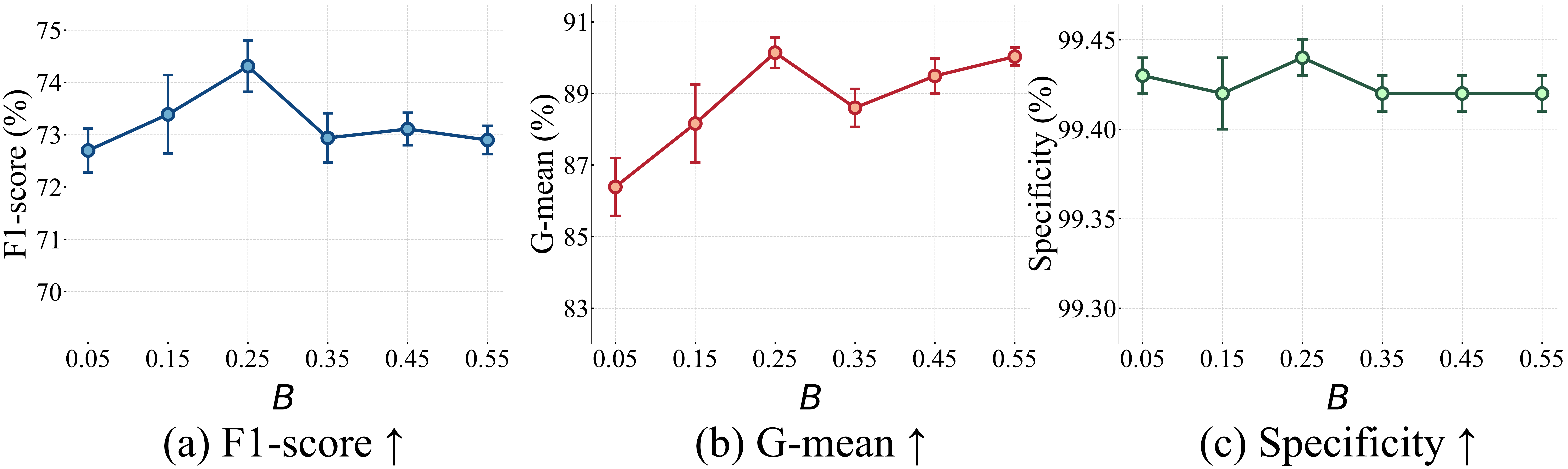}}
\caption{\textbf{The comprehensive performance of \texttt{\textbf{GDR-CIL}} with error bars by varying $B$ values on CIC-IDS2017 dataset (5 runs).}}
\label{CIC-IDS2017_different-adj-f1score-gmean-specificity-line}
\end{figure}

\textbf{Calibration Hyper-parameter $B$ Study.}
One hyper-parameter in the grouping mechanism is the calibration vector $\mathbf{v}_{\mathcal{G}}$, where $B$ controls its scale in $\mathbf{v}_{\mathcal{G}}=\left[\nicefrac{B}{\sqrt{n_1}},\dots,\nicefrac{B}{\sqrt{n_K}}\right]^{\intercal}$.
Here, we set $K=8$ and $\beta=0.005$, which has been proven to achieve the best performance in other experiments.
As revealed in Fig. \ref{CIC-IDS2017_different-adj-f1score-gmean-specificity-line}, \textit{F1-score} and \textit{G-mean} values rise with the increase of $B$ from 0.05 to 0.25, with highest results at $B=0.25$.
The \textit{Specificities} are close at each point with the best result at $B=0.25$.
Although \textit{G-mean} tends to increase after $B=0.35$, the overall best performance is at $B=0.25$.

\begin{figure}[tbp]
\centerline{\includegraphics[width=0.5\textwidth]{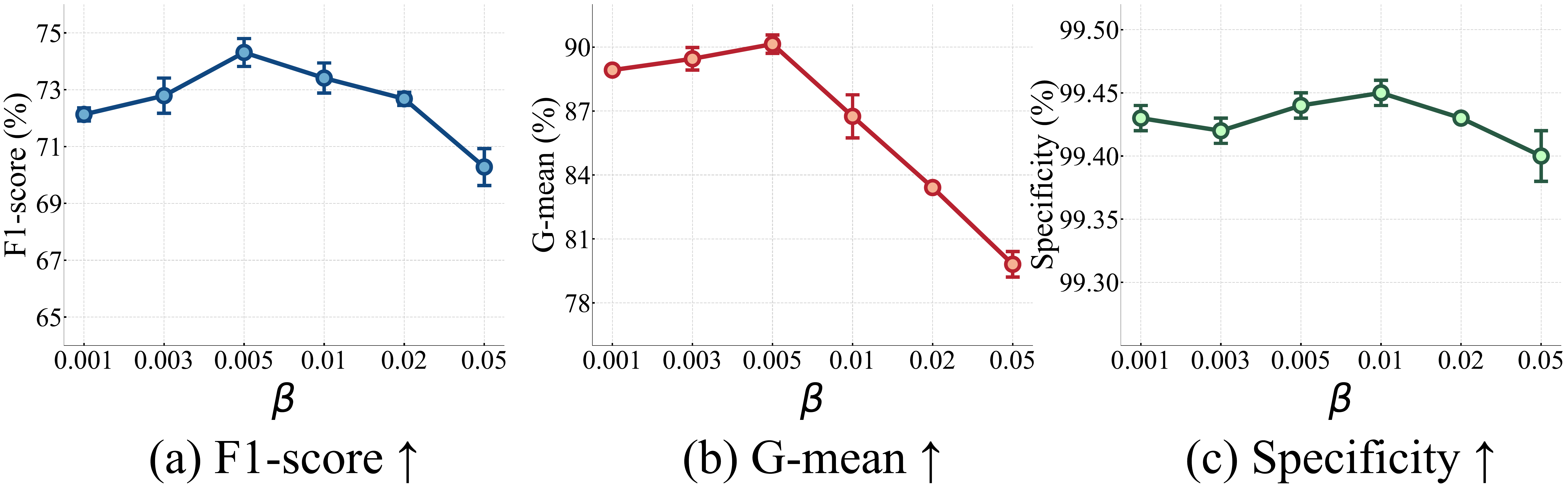}}
\caption{\textbf{The comprehensive performance of \texttt{\textbf{GDR-CIL}} with error bars by varying $\bm{\beta}$ values on CIC-IDS2017 dataset (5 runs).}}
\label{CIC-IDS2017_different-beta-f1score-gmean-specificity-line}
\end{figure}

\textbf{Temperature Hyper-parameter $\beta$ Study.}
Another relevant hyper-parameter $\beta$ is the temperature to control simplex values, and we report the corresponding results under different $\beta$ values in Fig. \ref{CIC-IDS2017_different-beta-f1score-gmean-specificity-line}.
Increasing the $\beta$ value from 0.001 to 0.005 encourages the loss value to assign more cost weights to the worst groups, consistently improving \textit{F1-score} and \textit{G-mean}.
Additionally, \textit{Specificity} achieves the highest value at $\beta=0.01$, close to the case when $\beta=0.005$.

\section{Conclusions and Limitations}\label{sec:conclusion}
\textbf{Empirical Findings.}
This work proposes a simple yet effective cost-sensitive approach \texttt{\textbf{GDR-CIL}} for imbalanced network traffic classification.
The Group \& Reweight strategy suppresses the class imbalance effect, and we translate the optimization process into a Stackelberg game.
Extensive results reveal the potential of our approach in risk-sensitive and class-imbalanced scenarios.

\textbf{Limitations and Future Extensions.}
Though our approach exhibits overall performance superiority with a clear interpretation, some additional configurations are still required in setups.
Here, we leave a more optimal grouping mechanism design as promising explorations in the future.

\section*{Acknowledgements}

This work is supported by the National Key Laboratory of Science and Technology on Blind Signal Processing under Grant No. 23007522, and the National Natural Science Foundation of China (NSFC) under Grant No. 62306326.



\bibliography{main_bib}
\bibliographystyle{icml2021}
\appendix

\onecolumn
\section{Benchmark Details}\label{Benchmarks}
CIC-IDS2017 \citep{sharafaldin2018toward} is a commonly used intrusion detection dataset with $15$ network traffic classes, including ``BENIGN'', which refers to normal, harmless network traffic, and $14$ kinds of the most up-to-date common attacks.
It provides CSV (Comma-Separated Values) files for machine and deep learning purposes, each containing $78$ different features, such as flow duration, maximum packet length, minimum packet length, number of forward packets, etc.

NSL-KDD \citep{tavallaee2009detailed} stands for ``NSL-KDD Data set for network-based Intrusion Detection Systems''.
It consists of four sub-datasets provided in CSV format: KDDTest+, KDDTest-21, KDDTrain+, KDDTrain+\_20Percent, where KDDTrain+\_20Percent and KDDTest-21 are subsets of KDDTrain+ and KDDTest+, respectively.
There are ``Normal'' and $4$ kinds of different types of attacks in these sets, including ``Denial of Service (DoS)'', ``Probe'', ``User-to-Root (U2R)'', and ``Remote to Local (R2L)'', as depicted in Fig. \ref{Data_quantity}b.
Each record in the dataset contains $41$ features.

UNSW-NB15 dataset \citep{moustafa2015unsw} is an intrusion evaluation dataset published by the University of New South Wales in 2015, with $10$ network traffic classes consisting of ``Normal'' and $9$ types of attacks.
It includes a training set (UNSW\_NB15\_training-set) and a testing set (UNSW\_NB15\_testing-set), both in CSV format, each set including $42$ features.


\section{Assumptions, Theorem and Proof}

\subsection{Assumptions}
We list all of the assumptions mentioned in this work.
These assumptions further serve the demonstration of Theorem \ref{them:cr} in the main paper.

\textbf{Assumption 1.} $\bm{\mathcal{L}}_{\mathcal{C}}(D;\bm\theta) $ is $\eta$-Lipschitz continuous \textit{w.r.t.} ${\bm \theta}$ for any dataset $D$.

\textbf{Assumption 2.}  There exist two constants $\xi > 0$ and $ 0<\gamma<1 $ such that $\Vert \nabla_{\bm \theta}{\mathcal{H}({\bm \omega}, \bm\theta)}\Vert_2 \le \xi $ and $\Vert \nabla_{\bm \omega}{\mathcal{H}({\bm \omega}, \bm\theta)}\Vert_2 \le \gamma $.

\subsection{Proof of Theorem \ref{them:cr}}\label{proof:cr}
{\textbf{Theorem \ref{them:cr}}[Convergence Rate for the Second Player]}
\textit{Let the iteration sequence in optimization be: $\cdots\mapsto\{{\bm \omega}^{(t-1)},{\bm \theta}^{(t)}\}\mapsto\{{\bm \omega}^{(t)},{\bm \theta}^{(t+1)}\}\mapsto\cdots\mapsto\{{\bm \omega}^*,{\bm \theta}^*\}$, with the converged equilibirum $({\bm \omega}^*,{\bm \theta}^*)$.
Under the Assumptions \ref{asum: loss}, \ref{asum: weight}, and suppose that $\Vert{\bf I} -\epsilon\nabla^{2}_{\bm \theta \bm \theta}\mathcal{F}({\bm \omega}^*;{\bm \theta}^*)\Vert_2<1-\epsilon\xi\eta$, we can have $\lim_{t\to\infty}\frac{\Vert{\bm \theta}^{(t+1)}-{\bm \theta}^*\Vert_2}{\Vert{\bm \theta}^{(t)}-{\bm \theta}^*\Vert_2}\leq 1$, and the iteration converges with the rate $\left(\Vert{\bf I} -\epsilon\nabla^{2}_{\bm \theta \bm \theta}\mathcal{F}({\bm \omega}^*;{\bm \theta}^*)\Vert_2+\epsilon\xi\eta\right)$.}

\textit{\textbf{Proof:}} 
Let the resulting stationary point be $[{\bm \omega}^*, {\bm \theta}^*]$, we denote the difference terms by $\hat{{\bm \omega}}={\bm \omega}-{\bm \omega}^*$ and $\hat{{\bm \theta}}={\bm \theta}-{\bm \theta}^*$.
Then, according to the optimization step, we can have the following equations:
\begin{align}\label{append_eq_sgd}
    {\bm \theta}^{(t+1)}={\bm \theta}^{(t)}-\epsilon\nabla_{{\bm \theta}}\mathcal{F}({\bm \omega}^{(t)};{\bm \theta}^{(t)})
    \implies
    \hat{\bm \theta}^{(t+1)}=\hat{\bm \theta}^{(t)}-\epsilon\nabla_{{\bm \theta}}\mathcal{F}({\bm \omega}^{(t)};{\bm \theta}^{(t)}).
\end{align}
Now we perform the first-order Taylor expansion of the ${\bm \theta}$ related function $\nabla_{{\bm \theta}}\mathcal{F}({\bm \omega}^{(t)};{\bm \theta})$ around ${\bm \theta}^*$ and can derive:
\begin{subequations}\label{append_eq_taylor}
	\begin{align}
		\nabla_{{\bm \theta}}\mathcal{F}({\bm \omega}^{(t)};{\bm \theta})
		&=
		\nabla_{{\bm \theta}}\mathcal{F}({\bm \omega}^{(t)};{\bm \theta}^*)
		+\nabla^{2}_{{\bm \theta}{\bm \theta}}\mathcal{F}({\bm \omega}^{(t)};{\bm \theta}^*)({\bm \theta}-{\bm \theta}^*)+{\mathcal{O}}(||{\bm \theta}-{\bm \theta}^*||)
		\\
		\nabla_{{\bm \theta}}\mathcal{F}({\bm \omega}^{(t)};{\bm \theta}^{(t)})
		&\simeq
		\nabla_{{\bm \theta}}\mathcal{F}({\bm \omega}^{(t)};{\bm \theta}^*)
		+\nabla^{2}_{{\bm \theta}{\bm \theta}}\mathcal{F}({\bm \omega}^{(t)};{\bm \theta}^*)({\bm \theta}^{(t)}-{\bm \theta}^*).
	\end{align}
\end{subequations}

Then we have the following result with the help of Assumptions \ref{asum: loss} and \ref{asum: weight}:
\begin{subequations}
	\begin{align}
		\Vert \nabla_{{\bm \theta}}\mathcal{F}({\bm \omega}^{(t)};{\bm \theta}^*) \Vert_2
		=&\Vert\nabla_{{\bm \theta}}\mathcal{F}({\bm \omega}^{(t)};{\bm \theta}^*)
		-\nabla_{{\bm \theta}}\mathcal{F}({\bm \omega}^*;{\bm \theta}^*) \Vert_2\\
        =&\Vert \mathbb{E}_{p(D)}\Big\{({\bm \omega}^{(t)} - {\bm \omega}^*)^{\intercal}\Big(\mathbf{\Gamma}\nabla_{{\bm \theta}}\bm{\mathcal{L}}_{\mathcal{C}}(D;\bm\theta^*) \Big)\Big\} \Vert_2 \\
        \underbrace{\le}_{\text{Jensen Inequality}}& \mathbb{E}_{p(D)}\Big\{ \Vert ({\bm \omega}^{(t)} - {\bm \omega}^*)^{\intercal}\Big(\mathbf{\Gamma}\nabla_{{\bm \theta}}\bm{\mathcal{L}}_{\mathcal{C}}(D;\bm\theta^*) \Big)\Vert_2\Big\} \\
        \le & \mathbb{E}_{p(D)}\Big\{ \Vert \nabla_{{\bm \theta}}\bm{\mathcal{L}}_{\mathcal{G}}(D;\bm\theta^*)\Vert_2  \Vert {\bm \omega}^{(t)} - {\bm \omega}^* \Vert_2 \Big\} \\
        < & \eta \mathbb{E}_{p(D)}\Big\{\Vert {\bm \omega}^{(t)} - {\bm \omega}^* \Vert_2 \Big\},
	\end{align} 
\end{subequations}
where the second inequality holds due to Assumption \ref{asum: loss}.

Below we focus on the term $ \Vert {\bm \omega}^{(t)} - {\bm \omega}^* \Vert_2 $.
Since
\begin{align}
    \tag*{}\Vert {\bm \omega}^{(t)} - {\bm \omega}^* \Vert_2 
    : = & \Vert \mathcal{H}({\bm \omega}^{(t-1)}, \bm\theta^{(t)}) - \mathcal{H}({\bm \omega}^*, \bm\theta^*) \Vert_2 \\
    \le & \Vert \mathcal{H}({\bm \omega}^{(t-1)}, \bm\theta^{(t)}) - \mathcal{H}({\bm \omega}^{(t-1)}, \bm\theta^*) \Vert_2 + \Vert \mathcal{H}({\bm \omega}^{(t-1)}, \bm\theta^*) - \mathcal{H}({\bm \omega}^*, \bm\theta^*) \Vert_2 \\
    \le &  \Vert \nabla_{\bm \theta}{\mathcal{H}({\bm \omega}^{(t-1)}, \bm\theta^*)}\Vert_2 \Vert {\bm \theta}^{(t)} - {\bm \theta}^* \Vert_2 + \Vert \nabla_{\bm \omega}{\mathcal{H}({\bm \omega}^*, \bm\theta^*)}\Vert_2 \Vert {\bm \omega}^{(t-1)} - {\bm \omega}^* \Vert_2 \\
    < & \xi \Vert {\bm \theta}^{(t)} - {\bm \theta}^* \Vert_2 + \gamma \Vert {\bm \omega}^{(t-1)} - {\bm \omega}^* \Vert_2 \\
    < & \cdots < \xi \sum_{i=0}^{t-1} \gamma^i\Big( \Vert {\bm \theta}^{(t-i)} - {\bm \theta}^* \Vert_2\Big) + \gamma^{t}  \Vert {\bm \omega}^{(0)} - {\bm \omega}^* \Vert_2\\
    = & \xi \Vert {\bm \theta}^{(t)} - {\bm \theta}^* \Vert_2 + \mathcal{O}(\gamma) \\
    \simeq & \xi \Vert {\bm \theta}^{(t)} - {\bm \theta}^* \Vert_2,
\end{align}
we can have 
\begin{align}\label{append_eq_mean_value}
    \Vert \nabla_{{\bm \theta}}\mathcal{F}({\bm \omega}^{(t)};{\bm \theta}^*) \Vert_2 < \xi \eta \Vert {\bm \theta}^{(t)} - {\bm \theta}^* \Vert_2
\end{align}

With \textbf{Eq.} \eqref{append_eq_sgd}, \textbf{Eq.} \eqref{append_eq_taylor} and \textbf{Eq.} \eqref{append_eq_mean_value}, we can derive the equation that:
\begin{subequations}
	\begin{align}
		\hat{\bm \theta}^{(t+1)}&=\hat{\bm \theta}^{(t)}-\epsilon\nabla_{\bm \theta}\mathcal{F}({\bm \omega}^{(t)};{\bm \theta}^{(t)})\\
		&=\hat{\bm \theta}^{(t)}-\epsilon\Big[\nabla_{\bm \theta}\mathcal{F}({\bm \omega}^{(t)};{\bm \theta}^*)
		+\nabla^{2}_{\bm \theta \bm \theta}\mathcal{F}({\bm \omega}^{(t)};{\bm \theta}^*)\hat{\bm \theta}^{(t)}\Big]\\
		&=\Big[{\bf I} -\epsilon \nabla^{2}_{\bm \theta \bm \theta}\mathcal{F}({\bm \omega}^{(t)};{\bm \theta}^*)\Big]\hat{\bm \theta}^{(t)} - \epsilon\nabla_{\bm \theta}\mathcal{F}({\bm \omega}^{(t)};{\bm \theta}^*) \\
		\implies
		\Vert\hat{\bm \theta}^{(t+1)}\Vert_2
		&\leq
		\Vert {\bf I} -\epsilon\nabla^{2}_{\bm \theta \bm \theta}\mathcal{F}({\bm \omega}^{(t)};{\bm \theta}^*)\Vert_2 \Vert \hat{\bm \theta}^{(t)}\Vert_2 + \epsilon \Vert \nabla_{\bm \theta}\mathcal{F}({\bm \omega}^{(t)};{\bm \theta}^*)\Vert_2 \\
		&\le \big( \Vert {\bf I} -\epsilon\nabla^{2}_{\bm \theta \bm \theta}\mathcal{F}({\bm \omega}^{(t)};{\bm \theta}^*)\Vert_2 + \epsilon \xi\eta\big)\Vert\hat{\bm \theta}^{(t)}\Vert_2
	\end{align}
\end{subequations}
Thus, when $ \Vert {\bf I} -\epsilon\nabla^{2}_{\bm \theta \bm \theta}\mathcal{F}({\bm \omega}^{(t)};{\bm \theta}^*)\Vert_2 < 1 - \epsilon \xi\eta $, we have 
\begin{subequations}
	\begin{align}
		\lim_{t\to\infty}\frac{\Vert\hat{\bm \theta}^{(t+1)}\Vert_2}{\Vert\hat{\bm \theta}^{(t)}\Vert_2}
		&\leq
		\lim_{t\to\infty}\Vert {\bf I} -\epsilon\nabla^{2}_{\bm \theta \bm \theta}\mathcal{F}({\bm \omega}^{(t)};{\bm \theta}^*)\Vert_2 + \epsilon \xi\eta
		\\
		&=\Vert {\bf I} -\epsilon\nabla^{2}_{\bm \theta \bm \theta}\mathcal{F}({\bm \omega}^*;{\bm \theta}^*)\Vert_2 + \epsilon \xi\eta \\
		&< 1.
	\end{align}
\end{subequations}

This completes the proof of Theorem \ref{them:cr}.
$\hfill\blacksquare$

\section{Baseline Details}\label{Baseline}

Here, we detail the mentioned baselines in this work as follows:
\begin{enumerate}
    \item Expected Risk Minimization (ERM) \citep{vapnik1991principles} loss: 
    ERM is a fundamental optimization framework in machine learning that learns model parameters by minimizing the average loss of the model over the training data.
    That is, it assigns the same weight to all classes.
    The commonly used ERM loss on classification tasks is the standard cross-entropy loss, defined for a single sample as $- \sum_{i=1}^{C} y_i \log(\hat{p}_i)$, where $C$ is the number of classes, $y_i$ is the true label, and $\hat{p}_i$ is the predicted probability.
    \item Focal loss \citep{lin2017focal}: 
    Focal loss is a variant of cross-entropy loss that focuses more on hard-to-classify examples, often used in tasks with class imbalance. It adds a factor $(1-\hat{p}_i)^\gamma$ to the standard cross-entropy criterion to down-weight the loss assigned to well-classified samples, where $\hat{p}_i$ represents the predicted probability and $\gamma \geq 0$ is the focusing parameter.
    In class-imbalance tasks, the weighting factor $\alpha_{i}$ is incorporated, which helps ensure that minority classes contribute more to the loss.
    Consequently, focal loss is expressed as $ - \alpha_{i}(1 -p_i)^\gamma \log(\hat{p}_i)$. 
    \item Class-Balanced (CB) \citep{cui2019class} loss: 
    CB loss introduces a balancing factor that adjusts the contribution of each class based on its effective number of samples.
    The effective number for class $i$ with $n_i$ samples is defined as $\nicefrac{(1-{\beta}^{n_i})}{(1-\beta)}$, where $\beta \in [0,1)$ controls the weighting, with values closer to $1$ giving more weight to the minority class.
    Thus, CB loss is formulated by $-\frac{1-\beta}{1-\beta^{n_i}}\cdot\log(\hat{p}_i)$.
    \item Mixup with Deferred Re-Weighting (Mixup-DRW) \citep{zhang2021bag} loss:
    Mixup-DRW combines the data augmentation from Mixup with a deferred re-weighting mechanism.
    Mixup constructs virtual examples as the linear interpolation of two given samples $(\mathbf{x}_i, y_i)$ and $(\mathbf{x}_j, y_j)$ from the training set, $\tilde{\mathbf{x}} = \lambda \mathbf{x}_i + (1 - \lambda) \mathbf{x}_j$ and $\tilde{y} = \lambda y_i + (1 - \lambda) y_j$, where $\lambda \in [0,1]$ is typically drawn from $\text{Beta}(\alpha, \alpha)$ with a hyperparameter $\alpha>0$.
    Mixup-DRW loss function can be written as $\sum_{i=1}^{N} \omega_{y_i} \cdot \lambda\bm{\mathcal{L}}(f(\tilde{\mathbf{x}}_i),y_i)+\omega_{y_j}\cdot(1-\lambda) \bm{\mathcal{L}}(f(\tilde{\mathbf{x}}_i),y_j)$, where $N$ is the mini-batch size, $\omega_{y_j}$ denotes the class weight of $y_i$, and $\bm{\mathcal{L}}(f(\tilde{\mathbf{x}}_i),y)$ is the loss (e.g., cross-entropy) between $\tilde{\mathbf{x}}_i$ and the actual label.
    In class-imbalance tasks, $\omega_{y_j}$ can be defined as $\nicefrac{1}{\text{frequcency}(y_i)}$.
    \item Remix with Deferred Re-Weighting (Remix-DRW) \citep{chou2020remix} loss:
    Remix-DRW builds upon the concept of Mixup-DRW.
    It mixes features of two samples in Mixup fashion and assigns the label that prefers to the minority class by providing a disproportionately higher weight to the minority class.
    The loss for Remix-DRW can be expressed as $\sum_{i=1}^{N} \omega_{y_i} \cdot \lambda^\prime\bm{\mathcal{L}}(f(\tilde{\mathbf{x}}_i),y_i)+\omega_{y_j}\cdot(1-\lambda^\prime) \bm{\mathcal{L}}(f(\tilde{\mathbf{x}}_i),y_j)$, where $\lambda^\prime$ is the Remix interpolation factor adjusted for class frequency.
    \item Label-Distribution-Aware Margin loss with Deferred Re-Weighting (LDAM-DRW) \citep{cao2019learning} loss:
    The LDAM loss is defined as $-\log\frac{e^{{f(\mathbf{x}_i)}_y-{\gamma}_y}}{\sum_{k} e^{{f(\mathbf{x}_i)}_k}}$, where ${\gamma}_y=\nicefrac{C}{n_y^\alpha}$ is the margin for class $y$, with $n_y$ being the number of samples in class $y$, $C$ a constant, and $\alpha\in[0,1]$ a hyperparameter that controls the strength of the margin based on class imbalance.
    Thus, the LDAM-DRW loss is given by $\sum_{i=1}^{N}\omega_y \cdot(-\log\frac{e^{{f(\mathbf{x})}_y-{\gamma}_y}}{\sum_{k} e^{{f(\mathbf{x})}_k}})$, where $\omega_y$ is the re-weighting factor that is inversely proportional to the class frequency, \textit{i.e.}, $\omega_y=\nicefrac{1}{\text{frequency}(y)}$.
    \item Label Distributionally Robust loss with Kullback-Leibler divergence (LDR-KL) \citep{zhu2023label} loss:
    With each class label assigned a distributional weight, LDR-KL formulates the loss in the worst case of the distributional weights regularized by KL divergence function.
    The LDR-KL loss can be written as $-\sum_{i=1}^{N}y_i \log q(y_i|\mathbf{x}_i) + \lambda\sum_{i=1}^{N}p(y_i) \log \frac{p(y_i)}{q(y_i|\mathbf{x}_i)}$, where $q(y_i|\mathbf{x}_i)$ is the model's predicted probability for the label $y_i$, $p(y_i)$ is the reference label distribution, and $\lambda$ is a weighting factor that balances the classification loss and the KL divergence term.
\end{enumerate}

To sum up, ERM forms the core of most supervised learning algorithms and is a natural baseline for comparison.
ERM is typically adapted by introducing cost-sensitive loss functions that adjust the penalties for incorrect predictions based on the class distribution or predefined costs.
For example, both focal loss and CB loss adjust weights based on class distribution. Mixup-DRW and Remix-DRW incorporate deferred re-weighting to adjust the importance of examples based on certain characteristics, making them forms of cost-sensitive learning.
LDAM-DRW introduces class-specific margins that vary depending on the class distribution and introduces DRW component that further enhances the cost-sensitive nature.
LDR-KL uses KL divergence to quantify the difference between the predicted and true label distributions, penalizing the model more for large discrepancies, which makes the method inherently cost-sensitive.

\section{Implementation Details}\label{Implementation}

\textbf{Training-Validation-Testing Dataset Splition.}
We describe the dataset-splitting strategy for all benchmarks.
For CIC-IDS2017 dataset, as shown in Fig. \ref{Data_quantity}a, 
the quantities of majority classes are highly greater than minority ones, thus we adopt an extraction strategy as the same in \citep{wang2022cttgan}.
We randomly extract $10,000$ traffic samples from the classes with data quantities exceeding $10,000$ while keeping the quantities of other classes unchanged.
Then we randomly split the dataset into training ($56\%$), validation ($14\%$), and testing ($30\%$) sets.
The training set consists of $91,784$ samples, the validation set $22,946$ samples, and the testing set the remaining $49,171$ samples.

For NSL-KDD dataset, we merge KDDTest-21 and KDDTrain+\_20Percent sets, introduced in Appendix \ref{Benchmarks}, into one set.
Then we perform the same training-validation-testing set split operation as CIC-IDS2017 dataset.
The training set consists of $20,743$ samples, the validation set $5,186$ samples, and the testing set the remaining $11,113$ samples.


For UNSW-NB15 dataset, we combine the training and testing sets, introduced in Appendix \ref{Benchmarks}, into one set.
Considering the high imbalance
among classes as shown in Fig. \ref{Data_quantity}c, we adopt the same extraction strategy as for CIC-IDS2017 dataset.
Then we use the same process of splitting the dataset into training, validation, and testing sets as CIC-IDS2017 dataset.
The training set consists of $37,346$ samples, the validation set $9,337$ samples, and the testing set the remaining $20,008$ samples.

The training set is to train the MLP,  the validation set is to tune hyper-parameters, such as the learning rate, and to determine when to stop training, and the testing set serves as an independent evaluation set to assess the generalization performance of the trained MLP.
Note that the testing set is not used for any training-related tasks to ensure unbiased evaluation.

\textbf{Training Details.}
Note that the number of groups $K$ is heuristically defined, and the class clustering results are based on the classification performance in the proxy training and the number of training samples in each class.
We treat the regularizer associated $B$ and the temperature $\beta$ as hyper-parameters.
For CIC-IDS2017 dataset, we set $K=8$, $B=0.25$, and $\beta=0.005$ based on the F1-score performance on the validation dataset. 
In the same way, we set $K=5$, $B=0.02$, $\beta=0.15$ in NSL-KDD dataset, and $K=10$, $B=0.02$, $\beta=0.1$ in UNSW-NB15 dataset given the validation dataset performance.

We provide the grouping results for each dataset.
In CIC-IDS2017 dataset, the fifteen classes are divided into $8$ groups.
The first group contains ``BENIGN''.
The second group includes ``DDoS'', ``DoS GoldenEye'', ``DoS Hulk'', and ``PortScan''.
The third group comprises ``SSH-Patator'', ``DoS Slowhttptest'', ``DoS slowloris'', and ``FTP-Patator''.
The fourth group combines ``Bot'' and ``Web Attack Brute Force''.
The remaining four classes represent four more groups.
The NSL-KDD dataset has $5$ groups, each representing a class.
Similarly, UNSW-NB15 dataset has $10$ groups.

To avoid overfitting, the MLP is augmented by the dropout module with a learning rate of $0.2$.
The parameters in the MLP neural network are initialized by a truncated normal distribution with a standard deviation of $3\mathrm{e}-2$.
The batch size is set to be $128$ as default.
In the training stage, we set up a learning rate scheduler that dynamically adjusts the learning rate based on the validation loss.
Concretely, if the validation loss does not improve for six epochs, the learning rate will be reduced by half.
The scheduler will not reduce the learning rate below $1\mathrm{e}-6$.
We implemented early stopping to halt training if validation loss stagnates for 30 consecutive epochs. The network is trained for a maximum of $200$ epochs.

\textbf{Python Implementation.}
This work's implementation is largely inspired by the group distributionally robust optimization method in \citep{sagawa2019distributionally}, where the GitHub link is (\url{https://github.com/kohpangwei/group_DRO}).
Further, we attach the loss function of \texttt{\textbf{GDR-CIL}} methods in Python as follows.

\begin{lstlisting}[language=Python, caption=Loss function in \texttt{\textbf{GDR-CIL}}]
import numpy as np
import torch

class LossComputer:
    def __init__(
                self, 
                criterion=nn.CrossEntropyLoss(reduction='none'), 
                n_groups: int, 
                group_counts: torch.Tensor, 
                B: float = 0.25,
                beta: float = 0.005, 
                normalize_loss: bool=False,
                device: torch.device,
            ):
        self.criterion = criterion
        self.device = device
        self.beta = beta
        self.normalize_loss = normalize_loss
        self.B = torch.tensor(B).to(device)
        self.n_groups = n_groups
        self.group_counts = group_counts.to(device)
        self.group_frac = self.group_counts / self.group_counts.sum()
        self.adv_probs = torch.ones(self.n_groups).cuda() / self.n_groups

    def loss(self, yhat, y, g=None):
        per_sample_losses = self.criterion(yhat, y)
        group_loss, group_count = self.compute_group_avg(per_sample_losses, g)
        actual_loss, weights = self.compute_robust_loss(group_loss, group_count)
        return actual_loss
    
    def compute_group_avg(self, losses, group_idx):
        group_map = (group_idx == torch.arange(self.n_groups).unsqueeze(1).long().cuda()).float()
        group_count = group_map.sum(1)
        group_denom = group_count + (group_count == 0).float()
        group_loss = (group_map @ losses.view(-1)) / group_denom
        return group_loss, group_count

    def compute_robust_loss(self, group_loss, group_count):
        adjusted_loss = group_loss
        if torch.all(self.adj > 0.0):
            adjusted_loss += self.adj / torch.sqrt(self.group_counts)
        if self.normalize_loss:
            adjusted_loss = adjusted_loss / (adjusted_loss.sum())
        self.adv_probs = self.adv_probs * torch.exp(self.step_size * adjusted_loss.data)
        self.adv_probs = self.adv_probs / (self.adv_probs.sum()) 
        robust_loss = group_loss @ self.adv_probs
        return robust_loss, self.adv_probs



\end{lstlisting}

\section{Platform and Computational Tools}

Throughout the paper, we use the Pytorch as the default toolkit to conduct all experiments.

\end{document}